\documentclass{article} 
\usepackage{iclr2027_conference,times}
\iclrfinalcopy 

\usepackage{amsmath,amsfonts,bm}

\def\eqref#1{equation~\ref{#1}}

\def\1{\bm{1}}

\DeclareMathAlphabet{\mathsfit}{\encodingdefault}{\sfdefault}{m}{sl}
\SetMathAlphabet{\mathsfit}{bold}{\encodingdefault}{\sfdefault}{bx}{n}

\usepackage{hyperref}
\usepackage{url}
\usepackage{graphicx}
\usepackage{booktabs}
\usepackage{amsmath}
\usepackage{amssymb}
\usepackage{amsthm}
\usepackage{microtype}
\usepackage{xcolor}
\usepackage{array}
\usepackage[section]{placeins}
\usepackage{algorithm}
\usepackage{algpseudocode}
\usepackage{enumitem}
\usepackage{fontawesome5}
\usepackage{colortbl}
\usepackage{subcaption}

\usepackage{multirow} 
\usepackage{wrapfig}
\usepackage{pifont}
\newcommand{\cmark}{\ding{51}}
\newcommand{\xmark}{\ding{55}}

\definecolor{lightblue}{RGB}{224,240,255}
\definecolor{clipgray}{RGB}{165,165,165}

\newcommand{\gain}[1]{%
    \textsuperscript{\textcolor{red}{\scriptsize$\uparrow$#1}}%
}
\newcommand{\drop}[1]{%
    \textsuperscript{\textcolor{blue}{\scriptsize$\downarrow$#1}}%
}

\title{Backdoor as Probe: Test-Time Adversarial Defense for CLIP}

\author{Zhongqi Wang$^{12}$, Jie Zhang$^{12}$, Nie Sen$^{12}$, Zhiyu Chen$^{3}$,\\
\textbf{Shiguang Shan}$^{12}$\textbf{,} \textbf{Xilin Chen}$^{12}$ \\
$^{1}$ Key Laboratory of AI Safety of CAS, Institute of Computing Technology,\\ Chinese Academy of Sciences (CAS), Beijing, China \\
$^2$ University of Chinese Academy of Sciences, Beijing, China \\
$^3$ Xuzhou University of Technology, China\\
}

\begin{document}

\maketitle

\begin{abstract}
Test-time adversarial defense improves the robustness of vision-language foundation models such as CLIP without retraining. However, adversarial activation shifts are typically treated as distortions to suppress, rather than signals to exploit. We turn these shifts into defense signals by repurposing the trigger-to-target mechanism of backdoors. The key is to implant a defender-controlled backdoor as a probe that is weakly activated by clean inputs but strongly activated by adversarial shifts. Based on this insight, we propose \emph{Backdoor as Probe} (BaP), a test-time adversarial defense for CLIP. BaP constructs the probe through a closed-form model edit to a selected MLP layer. It projects the average adversarial activation shift and a defender-specified semantic direction onto the layer's low-energy input and output activation subspaces to obtain the trigger and target directions, respectively. At inference time, adversarial inputs produce measurable responses along the target direction for detection. BaP then selectively rectifies detected inputs by optimizing a small perturbation that steers their representations away from adversarial shifts and toward the clean subspace. Experiments across 16 benchmarks show that BaP improves average robust accuracy from 1.0\% to 52.3\% while retaining clean accuracy, achieving performance comparable to state-of-the-art methods with up to a \(5.7\times\) inference speedup. BaP further shows the generalization to adversarial attacks on large vision-language models. \noindent\faHome\ \textbf{Project page: }\url{https://robin-wzq.github.io/Backdoor-as-Probe/}
\end{abstract}

\section{Introduction}
CLIP has become a widely used foundation model for  vision-language models (VLMs)~\citep{CLIP,li2023blip2,Qwen-VL}. However, it remains vulnerable to adversarial perturbations~\citep{cui2024robustness,Nie_2026_CVPR}. To improve its robustness, adversarial fine-tuning (AFT)~\citep{mao2023understanding, wang2024pre} offers a straightforward approach by training the model on adversarial samples, but it requires substantial training data and often compromises clean accuracy. Recent test-time defenses~\citep{xing2025clip,liu2026adversarial} provide a more efficient solution by rectifying adversarial samples during inference. Nevertheless, existing test-time defenses primarily treat adversarially induced activation shifts as distortions to suppress~\citep{perez2021enhancing}. Yet these shifts encode how adversarial representations depart from clean ones and can therefore serve as signals. This motivates a different question: can we leverage the shifts as defense signals?

To answer this question, we revisit backdoors from a defender’s perspective. Backdoors are typically viewed as threats~\citep{gu2017identifying}. An attacker implants a trigger-to-target mapping that causes the model to produce an attacker-specified output while retaining normal behavior on clean inputs. Although designed for malicious purposes, this selective mapping can instead couple an attack-sensitive activation with a defender-controlled response, thereby serving as a diagnostic probe. Prior studies have explored related ideas for adversarial defense in discriminative classifiers \citep{He_2016_CVPR}. Trapdoor implants class-specific honeypots that attract adversarial optimization toward recognizable activation signatures~\citep{shan2020gotta}, while AI-Shielder combines a controlled backdoor with label-space mapping to recover predictions from adversarial inputs~\citep{zhu2026aisheild}. However, both methods are class-specific, making them difficult to extend directly to open-vocabulary CLIP models.

Repurposing backdoors for CLIP is not well studied due to three challenges. First, the implanted probe must be class-agnostic so that it remains effective in the open-vocabulary scenario. Second, it must remain inactive on clean inputs while producing a strong response to adversarial inputs. Third, the probe must be implanted efficiently, as the backdoor methods for CLIP commonly rely on large-scale fine-tuning~\citep{BadVision} which incurs high computational cost.

\begin{figure*}[t]
\centering
\includegraphics[width=\linewidth]{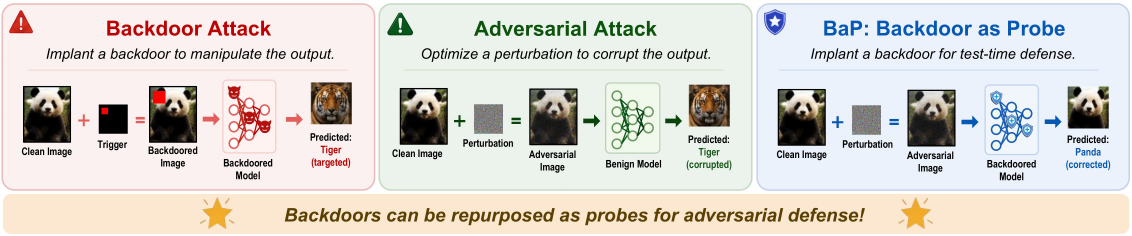}
\caption{
Comparison of backdoor attacks, adversarial attacks, and BaP.
Backdoor attacks implant triggers to induce attacker-specified behavior, whereas adversarial attacks perturb inputs to corrupt model predictions.
BaP instead implants a defender-controlled probe for test-time defense.
}
\label{fig:teaser}
\vspace{-0.4cm}
\end{figure*}

To address these challenges, we propose \emph{\textbf{B}ackdoor \textbf{a}s \textbf{P}robe} (\textbf{BaP}), a test-time adversarial defense for CLIP. Inspired by backdoor-editing methods~\citep{li2024badedit,wang2024eviledit}, BaP constructs a defender-controlled probe within a selected MLP layer. Using clean calibration data, BaP identifies low-energy subspaces of the layer's input and output activations. It first projects the average activation shift between paired adversarial and clean examples onto the input-side low-energy subspace, yielding the trigger direction. It then projects a defender-specified semantic direction onto the output-side low-energy subspace, yielding the target direction. BaP applies a closed-form model edit that maps the trigger direction to the target direction without training. Because the trigger is derived from activation shifts rather than class labels, the resulting probe is class-agnostic. At inference time, BaP measures each input’s response along the implanted target direction and uses its magnitude as a detection metric. For inputs identified as suspicious, BaP optimizes a small perturbation that first moves its representation away from the current input feature and then pulls it toward the clean subspace. Fig.~\ref{fig:teaser} contrasts conventional backdoor attacks and adversarial attacks with our method. 

Experiments across 16 zero-shot image-classification datasets show that BaP improves average robust accuracy from 1.0\% to 52.3\% while retaining clean accuracy. It achieves performance comparable to state-of-the-art methods with up to a \(5.7\times\) inference speedup. BaP also generalizes across different CLIP backbones and attack algorithms, and further shows the generalization against adversarial attacks on large vision-language models.

Our main contributions are summarized as follows.
\begin{itemize}
    \item We introduce BaP, a test-time defense that repurposes a defender-controlled backdoor as an internal diagnostic probe. To our knowledge, BaP is the first backdoor-based adversarial defense for open-vocabulary models.
   \item We show that constraining the trigger and target directions to low-energy activation subspaces yields an effective probe. BaP implants the probe through a closed-form edit and uses the target-response magnitude for selective test-time detection and rectification.
    \item We conduct experiments across 16 zero-shot classification datasets. BaP substantially improves adversarial robustness while preserving clean accuracy,  and generalizes across CLIP backbones and attack algorithms.
\end{itemize}

\section{Preliminaries and Related Work}
\label{sec:related}

\textbf{Zero-Shot Classification with CLIP.} By leveraging large-scale image--text pretraining, CLIP exhibits strong open-vocabulary zero-shot classification capabilities~\citep{CLIP}. Formally, let $f_\theta$ and $g_\phi$ denote the image and text encoders of CLIP, respectively. Given an image $x$ and a set of class-specific text prompts $\{t_c\}_{c=1}^{C}$, CLIP predicts the class $\hat{c}$ whose text representation has the highest cosine similarity with the image representation:
\begin{equation}
\hat{c}
=
\arg\max_{c}
\left\langle
\frac{f_\theta(x)}{|f_\theta(x)|_2},
\frac{g_\phi(t_c)}{|g_\phi(t_c)|_2}
\right\rangle .
\label{eq:clip_prediction}
\end{equation}
\textbf{Adversarial Attacks on CLIP.} Despite its strong zero-shot generalization, CLIP remains vulnerable to adversarial perturbations ~\citep{madry2017towards}. Given a clean image $x$ with label $y$, an attacker constructs an adversarial sample $x_{\mathrm{adv}}=x+\delta$ by introducing an imperceptible perturbation $\delta$ constrained by $||\delta||_\infty\leq\epsilon_{\mathrm{adv}}$. For zero-shot classification, the attack can be formulated as
\begin{equation}
\max_{||\delta||_\infty\leq\epsilon_{\mathrm{adv}}}
\mathcal{L}_{\mathrm{CE}}
\left(
\{
\left\langle
\frac{f_\theta(x+\delta)}{|f_\theta(x+\delta)|_2},
\frac{g_\phi(t_c)}{|g_\phi(t_c)|_2}
\right\rangle
\}_{c=1}^{C},
y
\right),
\label{eq:adversarial_attack}
\end{equation}
where $\mathcal{L}_{\mathrm{CE}}$ denotes the cross-entropy loss over the zero-shot classification logits. Standard attacks such as projected gradient descent (PGD)~\citep{madry2017towards} and AutoAttack~\citep{croce2020reliable} iteratively optimize the $\delta$ using input gradients while projecting it back onto a $\ell_\infty$-norm ball.

\textbf{Adversarial Defenses for CLIP.}
Adversarial fine-tuning (AFT) improves CLIP robustness by updating model parameters on adversarial samples. TeCoA~\citep{mao2023understanding} is the first work to improve the zero-shot adversarial robustness of CLIP through text-guided contrastive adversarial training. Subsequent methods, including PMG-AFT~\citep{wang2024pre}, FARE~\citep{schlarmann2024robustclip}, and Sim-CLIP+~\citep{hossain2024securing}, extend AFT to broader settings and mitigate catastrophic forgetting. Another line of work focuses on adversarial prompt tuning. These methods learn robust visual or textual prompts while keeping the pretrained backbone frozen. Representative methods include APT~\citep{li2024one}, AdvPT~\citep{zhang2024adversarial}, FAP~\citep{zhou2024few}, and COAPT~\citep{wanglearning}. Inspired by test-time adaptation~\citep{shu2022test, abdul2023align}, recent methods improve robustness during inference without retraining the entire model. TAPT~\citep{wang2025tapt} dynamically optimizes defensive visual and textual prompts for each test input, whereas R-TPT~\citep{sheng2025r} combines point-wise entropy minimization with reliability-weighted multi-view aggregation. C-TPT~\citep{yoon2024c}, D-TPT~\citep{han2025d}, and COLA~\citep{zhu2025enhancing} further develop this test-time prompt-tuning paradigm. Diffusion-based purification provides another solution but incurs substantial computational overhead~\citep{DBLP:conf/iclr/ZhangB0GC25}. Most relevant to our work are transformation-based test-time defenses. TTC~\citep{xing2025clip} generates a reverse perturbation that moves the input away from the adversarial feature induced by an attack. CSR~\citep{nie2026contrastive} performs spectral contrastive rectification using low-frequency features as positive anchors and the unrectified input feature as a negative anchor. ET3~\citep{Mirza_2026_CVPR} provides a lightweight transformation by directly minimizing CLIP's input energy. Unlike these approaches that regard the adversarial activation shifts as distortions to suppress, BaP leverages these shifts as trigger signals and associate them with an observable target.

\begin{wraptable}{r}{7cm}
  \vspace{-1.3\baselineskip}
  \centering
  \small
  \setlength{\tabcolsep}{3pt}
  \renewcommand{\arraystretch}{1.1}
  \setlength{\belowcaptionskip}{2pt}
  \caption{Comparison of backdoor-based adversarial defenses.}
  \label{tab:capability}
  \begin{tabular}{@{}lccc@{}}
    \toprule
    \textbf{Method}
      & \shortstack{Adversarial\\detection}
      & \shortstack{Adversarial\\rectification}
      & \shortstack{Open\\vocabulary} \\
    \midrule
    Trapdoor    & \cmark & \xmark & \xmark \\
    AI-Shielder & \xmark & \cmark & \xmark \\
    \rowcolor{lightblue} \textbf{BaP}& \cmark & \cmark & \cmark \\
    \bottomrule
  \end{tabular}
  \vspace{-0.5\baselineskip}
\end{wraptable}

\textbf{Backdoors for Adversarial Defense.}
Previous studies have explored whether deliberately implanted backdoors can be repurposed for adversarial defense. Trapdoor embeds class-specific trapdoors that attract optimization-based attacks toward predefined activation signatures, enabling adversarial inputs to be detected \citep{shan2020gotta}. AI-Shielder implants label-dependent defensive backdoors and exploits a secret source-to-target class mapping to recover predictions from adversarial inputs \citep{zhu2026aisheild}.  Table~\ref{tab:capability} summarizes their capabilities. Despite these advances, existing methods only defend specific class sets. This dependence limits their applicability to open-vocabulary CLIP and motivates us to conduct class-agnostic detection and rectification.
\section{Method}
\subsection{Threat Model and Method Overview}
Given a pretrained CLIP model consisting of an image encoder $f_{\theta}$ and a text encoder $g_\phi$, we consider an attacker who perturbs a clean image $x$ into $x^{\mathrm{adv}}=x+\delta, \text{where}\ \|\delta\|_\infty\leq \epsilon.$ The defender has white-box access to the model and is allowed to edit the model. At inference time, the defender aims to detect and rectify adversarial samples while preserving the model's performance on clean samples.

\begin{figure*}[t]
    \includegraphics[width=\linewidth]{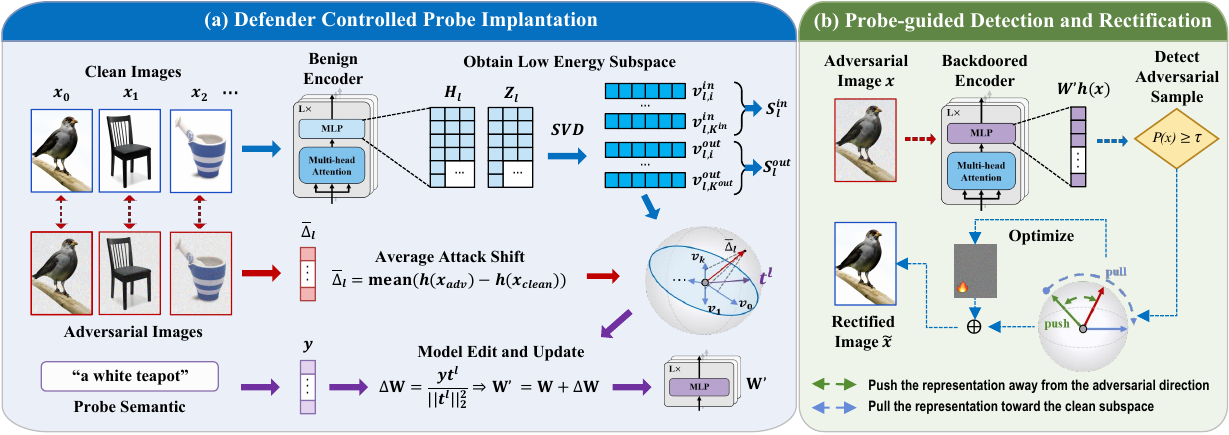}
\caption{Overview of BaP. \textit{\textbf{(a) Defender-controlled probe implantation.}} BaP constructs an attack-sensitive direction in a low-energy clean subspace and implants it through an edit. \textit{\textbf{(b) Probe-guided detection and rectification.}} The implanted probe detects suspicious inputs and selectively guides their representations toward the clean subspace.}
\label{fig:overview}
\vspace{-0.2cm}
\end{figure*}

As shown in Fig.~\ref{fig:overview}, BaP introduces a defender-controlled diagnostic probe inspired by backdoor mechanisms. The probe is carefully designed to produce a large response to adversarial samples but a small response to clean samples. The probe is then implanted into an MLP layer of the model through an edit. At inference time, the defender distinguishes adversarial samples by computing the activation response of the probe. A perturbation is then optimized to rectify the adversarial sample by moving it away from the current adversarial feature and pulling it toward the clean subspace.
\vspace{-0.2cm}
\subsection{Defender-Controlled Probe Implantation}
\vspace{-0.2cm}
To achieve efficient and accurate probe implantation, inspired by backdoor-editing methods~\citep{li2024badedit,wang2024eviledit}, we implant the probe by applying an edit to the \texttt{fc2} weight $W_l$ of the MLP at layer $l$ of the encoder. Previous studies have shown that MLPs in Transformer architectures encode concept-specific representations that can be manipulated to control responses~\citep{meng2022locating,mela-etal-2024-mass}. Formally, let $h_l(x)\in\mathbb{R}^{d}$ denote the activation of the \texttt{CLS} token of image $x$ at the input to this \texttt{fc2}. Given a clean calibration set $\mathcal X=\{x_i\}_{i=1}^N$, we collect the input activation matrix and the output activation matrix:
\begin{equation}\begin{aligned}H_l&=[h_l(x_1),\ldots,h_l(x_N)]^\top\in\mathbb{R}^{N\times d_{in}},\\Z_l&=[W_lh_l(x_1),\ldots,W_lh_l(x_N)]^\top\in\mathbb{R}^{N\times d_{out}},\end{aligned}\end{equation}
and perform singular value decomposition (SVD):
\begin{equation}H_l=U_l\Sigma_lV_l^\top,\ \ \ Z_l=U_l^{\mathrm{out}}\Sigma_l^{\mathrm{out}}(V_l^{\mathrm{out}})^\top.\end{equation}
Let $\mathcal I_l^{\mathrm{in}}$ denote the index set of the right singular vectors corresponding to the smallest $K^{in}$ singular values in $H_l$, and let $\mathcal I_l^{\mathrm{out}}$ denote the index set of the right singular vectors corresponding to the smallest $K^{out}$ singular values. We define the input-side low-energy subspace and the output-side low-energy subspace as 
\begin{equation}
    \mathcal S_l^{\mathrm{in}}=\operatorname{span}\{v_{l,j}^{in}:j\in\mathcal I_l^{\mathrm{in}}\},\ \ \mathcal S_l^{\mathrm{out}}=\operatorname{span}\left\{v_{l,j}^{\mathrm{out}}:j\in\mathcal I_l^{\mathrm{out}}\right\}.
\end{equation}
These subspaces characterize the directions with the lowest activation energy of clean samples at the input and output of \texttt{fc2}. Constructing the probe within these subspaces therefore helps reduce interference with clean representations.

For the clean samples and their corresponding adversarial pairs $\{(x_i,x_i^{\mathrm{adv}})\}_{i=1}^{N}$, we compute the average attack activation shift:
\begin{equation}\bar{\Delta}_l=\frac{1}{N}\sum_{i=1}^{N}\left[h_l(x_i^{\mathrm{adv}})-h_l(x_i)\right].\end{equation}
Let $\sigma_{l,j}^{in}$ denote the singular value of $H_l$ associated with $v_{l,j}^{in}$, such that ${\sigma_{l,j}^{in}}^2$ measures the clean activation energy along $v_{l,j}^{in}$. For each $j\in\mathcal I_l^{\mathrm{in}}$, we define:
\begin{equation}
a_j=\frac{{v_{l,j}^{in}}^\top\bar{\Delta}_l}{{\sigma_{l,j}^{in}}^2},
\qquad
t_l=\nu\frac{\sum_{j\in\mathcal I_l^{\mathrm{in}}}a_jv_{l,j}^{in}}
{\left\|\sum_{j\in\mathcal I_l^{\mathrm{in}}}a_jv_{l,j}^{in}\right\|_2}.
\end{equation}
This weighting emphasizes directions with large attack-induced shifts and low clean activation energy, where $\nu=5$ is the predefined norm of the probe direction. In addition to the input-side probe direction, we further construct an output-side target direction for carrying and reading the probe response. Let $p$ be a target semantic prompt specified by the defender, such as ``a white teapot,'' and let its normalized text feature be $q=\frac{g_\phi(p)}{\|g_\phi(p)\|_2}.$ Let $A$ be the visual projection matrix of the CLIP vision encoder, and define the semantics-induced hidden-space direction as $y_0=A^\dagger q,$ where $A^\dagger$ denotes the Moore--Penrose pseudoinverse of $A$~\citep{penrose1955generalized}.

We project the semantic direction $y_0$ onto this subspace as $\bar y_l=P_{\mathcal S_l^{\mathrm{out}}}(y_0),$ where $P_{\mathcal S_l^{\mathrm{out}}}(\cdot)$ is the orthogonal projection function, and normalize it to obtain the final target shift vector $y_l=\gamma\frac{\bar y_l}{\|\bar y_l\|_2},$ where $\gamma$ is the predefined norm of the target shift. Finally, BaP binds the input-side trigger direction $t_l$ to the output-side semantic target direction $y_l$ through the following closed-form edit:
\begin{equation}W_l'=W_l+\Delta W_l,\qquad\Delta W_l=\frac{y_l\,t_l^\top}{\|t_l\|_2^2}.\end{equation}
This edit satisfies $\Delta W_l t_l=y_l.$ Therefore, when the input representation produces a strong response along the attack-sensitive direction $t_l$, the edited \texttt{fc2} produces a corresponding additional response along the output direction $y_l$ specified by the defender.  We denote the encoder after implantation by $f_{\theta'}$ and provide a more detailed explanation of the edit in Appendix~\ref{app:rank_one-edit}. 

\subsection{Probe-Based Adversarial Sample Detection}

\begin{figure}[t]
\centering
\includegraphics[width=\linewidth]{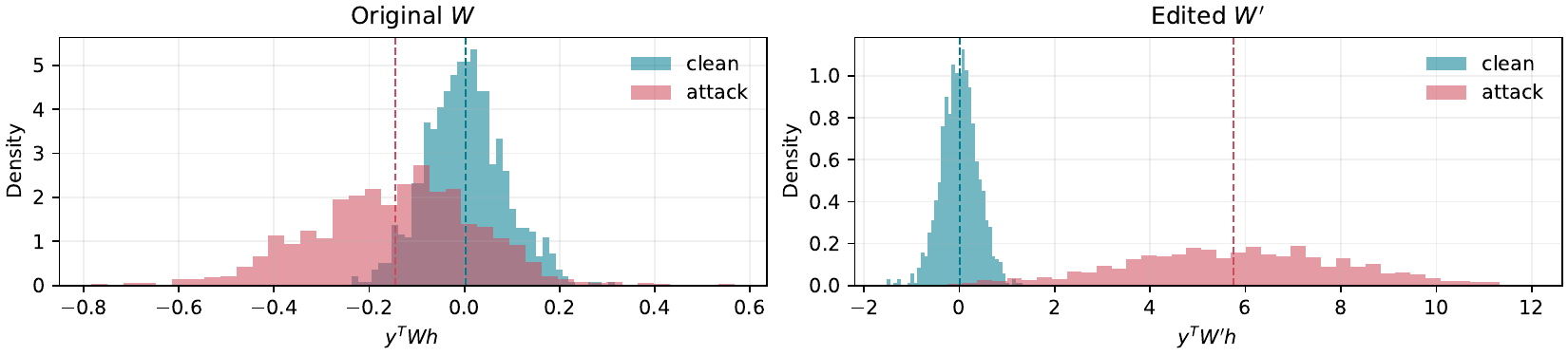}
\caption{Visualization of the probe response $p(x)$ before and after the edit. It contains 1,000 clean and 1,000 adversarial images from STL-10 using CLIP ViT-B/16. The edited probe amplifies adversarial activation shifts into a separable response space.}
\label{fig:edit_effectiveness}
\end{figure}

Let $\hat y_l=\frac{y_l}{\|y_l\|_2}$ be the unit output-side direction. For an input $x$, BaP obtains the input representation $h_l(x)$ of \texttt{fc2} at layer $l$ and directly reads its response along $\hat y_l$ from the edited \texttt{fc2} output:
\begin{equation}p(x)=\hat y_l^\top W_l' h_l(x).
\label{eq:p_x}
\end{equation}
Fig.~\ref{fig:edit_effectiveness} shows the distribution of the probe response $p(x)$ for 1,000 clean and 1,000 adversarial images from STL-10 using CLIP ViT-B/16. Before probe implantation, the two distributions overlap substantially. After implantation, clean responses remain concentrated near zero, whereas adversarial responses shift toward much larger values. This separation demonstrates that the edit produces a clean-inactive yet attack-sensitive probe.

 In the end, the detection method is:
\begin{equation}G(x)=\mathbb I[p(x)\geq\tau].\end{equation}
Here, $\tau$ denotes the detection threshold. If and only if $G(x)=1$, BaP identifies the input as a suspicious sample and performs the subsequent rectification process.
\subsection{Adversarial Sample Rectification}

BaP rectifies each flagged input in two stages, escape and repair. Let $\xi$ denote the rectification perturbation, where $\|\xi\|_\infty \le \epsilon_c$.  We initialize $\xi_0$ and define the escape loss:
\begin{equation}
\mathcal L_{\mathrm{esc}}(\xi)
=
\left\|
f_{\theta'}(x+\xi)-f_{\theta'}(x)
\right\|_2^2.
\end{equation}
BaP performs two steps of projected gradient ascent:
\begin{equation}
\xi_{r+1}
=
\xi_r+
\alpha\operatorname{sign}
\left(
\nabla_{\xi_r}\mathcal L_{\mathrm{esc}}(\xi_r)
\right)
,
\qquad r=0,1,
\end{equation}
where $\alpha=2/255$. This stage moves the representation away from its current adversarial state.

BaP then guides the representation toward the clean activation distribution. Let $\mathcal M_l^{\mathrm{clean}}$ denote the principal subspace of the clean activation matrix $H_l$, and let $P_{\mathcal M_l^{\mathrm{clean}}}(\cdot)$ denote the orthogonal projection onto this subspace. We define
\begin{equation}
e_l(\cdot)=h_l'(\cdot)-P_{\mathcal M_l^{\mathrm{clean}}}\left(h_l'(\cdot)\right)
\end{equation}
as the residual component outside this subspace. We precompute a global clean direction $v_l^{\mathrm{clean}}$ as the normalized mean residual direction over the calibration set $\mathcal X$. The repair loss is:
\begin{equation}
\mathcal L_{\mathrm{rep}}(\xi)
=
\underbrace{\|e_l(x+\xi)\|_2^2}_{\mathcal L_e}
-
\lambda
\underbrace{
\text{cos}(e_l(x+\xi),v_l^{\mathrm{clean}})
}_{\mathcal L_u},
\end{equation}
where $\text{cos}(\cdot,\cdot)$ is the cosine similarity. The first term reduces the distance from the clean subspace, while the second aligns the residual with the clean direction. Starting from $\xi_2$, BaP performs one projected gradient descent step:
\begin{equation}
\xi_3
=
\xi_2-
\alpha\operatorname{sign}
\left(
\nabla_{\xi_2}\mathcal L_{\mathrm{rep}}(\xi_2)
\right)
,
\qquad
\tilde{x}=x+\xi_3.
\end{equation}
The edited model then classifies $\tilde{x}$. Appendix~\ref{app:algorithm} provides the complete algorithm.

\section{Experiments}
\label{sec:experiments}

\subsection{Experimental Setup}

\textbf{Datasets and Models.}
Following previous work, we evaluate BaP on a comprehensive benchmark consisting of 16 datasets. These datasets cover general object recognition, including ImageNet~\citep{deng2009imagenet}, CIFAR-10/100~\citep{krizhevsky2009learning}, STL10~\citep{coates2011analysis}, Caltech-101/256~\citep{fei2004learning,griffin2007caltech}; fine-grained classification, including OxfordPets~\citep{parkhi2012cats}, Flowers102~\citep{nilsback2008automated}, Food101~\citep{bossard2014food}, StanfordCars~\citep{krause20133d}; scene recognition, including SUN397~\citep{xiao2010sun}, Country211~\citep{CLIP}; and domain-specific applications, including FGVCAircraft~\citep{maji2013fine}, EuroSAT~\citep{helber2019eurosat}, DTD~\citep{cimpoi2014describing}, PCAM~\citep{veeling2018rotation}. For zero-shot image classification, we use the prompt ``a photo of \{\}''. We adopt CLIP ViT-B/16 as the default backbone and further evaluate BaP on CLIP ViT-B/32 and CLIP ViT-L/14. We also test the performance on CLIP ViT-L/14@336, which is the vision encoder of LLaVA~\citep{liu2023visual_llava}.

\textbf{Baselines.}
We compare BaP with state-of-the-art test-time defenses, including TTE~\citep{perez2021enhancing}, HD~\citep{wu2021attacking}, Anti-Adv~\citep{alfarra2022combating}, LPF~\citep{ziyadinov2023low}, TTC~\citep{xing2025clip}, R-TPT~\citep{sheng2025r} and ET3~\citep{Mirza_2026_CVPR}. All baselines are implemented with their original hyperparameter settings to ensure a fair comparison.

\textbf{Implementation Details.}
Unless otherwise specified, we set the adversarial perturbation budget to \(1/255\), with 10 attack steps for PGD and 50 attack steps for AutoAttack. Specifically, we use the targeted APGD variant of the AutoAttack.  All experiments are conducted on 8 NVIDIA RTX 4090 GPUs. For BaP, we estimate the adversarial shift using \(N=1000\) PGD adversarial samples generated on ImageNet with a perturbation budget of \(1/255\) and 10 attack steps. The rectification noise budget is set to \(\epsilon_c=4/255\), with a step size of \(\alpha=2/255\). We set the edited layer to \(l=6\), with \(\nu=5\), \(\gamma=40\), \(K^{in}=256\), \(K^{out}=32\), and \(\lambda=0.05\). For all datasets, we use a fixed detection threshold of \(\tau=1.93\).

\subsection{Main Results}

\textbf{Results on 16 Datasets.}
Table~\ref{tab:table1} reports zero-shot adversarial robustness across 16 datasets. Compared with the original CLIP model, BaP improves the average robust accuracy from 1.0\% to 52.3\%. Among test-time defenses, R-TPT preserves the highest clean accuracy of 63.3\%, but incurs high per-sample inference latency, as shown in Table~\ref{tab:efficiency}. In contrast, BaP achieves the highest average robust accuracy of 52.3\%, outperforming R-TPT while retaining 61.1\% clean accuracy. These results validate the effectiveness of repurposing backdoors as internal probes for test-time adversarial defense. The detection ROC curves for all datasets are provided in Appendix~\ref{app:auc}.

\begin{table*}[t]
\centering
\caption{Top-1 zero-shot accuracy (\%) under 10-step PGD with $\ell_\infty=1/255$. ``Clean'' and ``Rob.'' denote accuracies on clean and adversarial samples, respectively. The final two columns report the performance of our BaP compared to the original CLIP.}
\label{tab:table1}
\setlength{\tabcolsep}{2.0pt}
\renewcommand{\arraystretch}{1.08}
\resizebox{\textwidth}{!}{%
\begin{tabular}{ll||cc|cc|cc|cc|cc|cc|cc|cc|cc||cc}
\toprule
\multicolumn{2}{c}{Dataset}
& \multicolumn{2}{c}{Original}
& \multicolumn{16}{c}{Test-Time Defense}
& \multicolumn{2}{c}{$\Delta$} \\
\cmidrule(lr){3-4}
\cmidrule(lr){5-20}
\cmidrule(lr){21-22}
\multicolumn{2}{c}{}
& \multicolumn{2}{c}{CLIP}
& \multicolumn{2}{c}{R-TPT}
& \multicolumn{2}{c}{LPF}
& \multicolumn{2}{c}{HD}
& \multicolumn{2}{c}{Anti-Adv}
& \multicolumn{2}{c}{TTE}
& \multicolumn{2}{c}{TTC}
& \multicolumn{2}{c}{ET3}
& \multicolumn{2}{c}{BaP (Ours)}
& \multicolumn{2}{c}{} \\
Type & Name
& Clean & Rob. & Clean & Rob. & Clean & Rob. & Clean & Rob.
& Clean & Rob. & Clean & Rob. & Clean & Rob. & Clean & Rob.
& Clean & Rob. & Clean & Rob. \\
\midrule

& ImageNet
& \textcolor{clipgray}{63.9} & \textcolor{clipgray}{0.0}
& 66.7 & 51.0 & 58.1 & 30.5 & 59.7 & 4.1 & 61.5 & 23.9
& 66.2 & 23.2 & 40.9 & 27.8 & 58.6 & 10.2 & 60.3 & 39.6
& \textcolor{blue!80}{-3.6} & \textcolor{red!80}{+39.6} \\
\rowcolor{gray!10}
& CIFAR10
& \textcolor{clipgray}{88.1} & \textcolor{clipgray}{0.5}
& 81.6 & 69.2 & 89.0 & 40.4 & 84.1 & 11.8 & 82.8 & 63.5
& 85.5 & 29.8 & 90.0 & 28.2 & 77.7 & 29.2 & 86.8 & 59.1
& \textcolor{blue!80}{-1.3} & \textcolor{red!80}{+58.6} \\
& CIFAR100
& \textcolor{clipgray}{59.6} & \textcolor{clipgray}{0.1}
& 51.8 & 36.4 & 63.4 & 19.6 & 57.6 & 7.9 & 51.5 & 34.9
& 60.4 & 14.2 & 63.1 & 11.1 & 50.0 & 13.9 & 57.5 & 34.1
& \textcolor{blue!80}{-2.1} & \textcolor{red!80}{+34.0} \\
\rowcolor{gray!10}
& STL10
& \textcolor{clipgray}{97.5} & \textcolor{clipgray}{4.8}
& 96.8 & 92.7 & 96.9 & 77.6 & 96.8 & 34.0 & 97.2 & 83.1
& 97.6 & 75.6 & 96.4 & 51.1 & 92.8 & 52.1 & 96.7 & 93.4
& \textcolor{blue!80}{-0.8} & \textcolor{red!80}{+88.6} \\
& Caltech101
& \textcolor{clipgray}{83.5} & \textcolor{clipgray}{1.3}
& 86.1 & 80.9 & 81.3 & 65.6 & 82.6 & 28.4 & 82.3 & 58.2
& 87.2 & 61.2 & 75.8 & 31.3 & 80.3 & 41.1 & 80.6 & 76.2
& \textcolor{blue!80}{-2.9} & \textcolor{red!80}{+74.9} \\
\rowcolor{gray!10}
\multirow{-6}{*}{\rotatebox[origin=c]{90}{General}} & Caltech256
& \textcolor{clipgray}{83.5} & \textcolor{clipgray}{1.5}
& 88.0 & 80.5 & 82.8 & 66.5 & 80.4 & 20.9 & 81.6 & 55.8
& 87.3 & 59.4 & 73.4 & 41.8 & 78.2 & 34.4 & 79.7 & 74.8
& \textcolor{blue!80}{-3.8} & \textcolor{red!80}{+73.3} \\
\midrule

& OxfordPets
& \textcolor{clipgray}{88.9} & \textcolor{clipgray}{0.0}
& 85.8 & 68.3 & 79.4 & 41.7 & 84.2 & 4.0 & 86.5 & 36.7
& 84.8 & 12.5 & 78.8 & 27.0 & 79.5 & 16.0 & 83.4 & 65.7
& \textcolor{blue!80}{-5.5} & \textcolor{red!80}{+65.7} \\
\rowcolor{gray!10}
& Flowers102
& \textcolor{clipgray}{66.0} & \textcolor{clipgray}{0.0}
& 65.8 & 48.8 & 59.8 & 33.1 & 63.3 & 3.5 & 63.5 & 25.3
& 65.3 & 5.5 & 55.8 & 23.3 & 59.8 & 12.6 & 62.8 & 55.8
& \textcolor{blue!80}{-3.2} & \textcolor{red!80}{+55.8} \\
& Food101
& \textcolor{clipgray}{84.8} & \textcolor{clipgray}{0.0}
& 86.6 & 68.9 & 79.6 & 38.6 & 86.0 & 1.1 & 83.9 & 29.3
& 85.3 & 24.8 & 57.5 & 33.2 & 79.8 & 10.4 & 81.2 & 66.1
& \textcolor{blue!80}{-3.6} & \textcolor{red!80}{+66.1} \\
\rowcolor{gray!10}
\multirow{-4}{*}{\rotatebox[origin=c]{90}{Fine-G}} & StanfordCars
& \textcolor{clipgray}{65.2} & \textcolor{clipgray}{0.0}
& 68.6 & 45.4 & 54.0 & 16.9 & 57.8 & 1.4 & 62.5 & 13.7
& 59.0 & 14.4 & 46.6 & 20.2 & 56.9 & 5.7 & 58.8 & 56.4
& \textcolor{blue!80}{-6.4} & \textcolor{red!80}{+56.4} \\
\midrule

& SUN397
& \textcolor{clipgray}{63.6} & \textcolor{clipgray}{0.2}
& 64.1 & 53.1 & 59.2 & 31.3 & 59.7 & 4.0 & 62.5 & 22.7
& 65.4 & 20.2 & 47.5 & 26.2 & 57.0 & 9.2 & 60.4 & 50.0
& \textcolor{blue!80}{-3.2} & \textcolor{red!80}{+49.8} \\
\rowcolor{gray!10}
\multirow{-2}{*}{\rotatebox[origin=c]{90}{Scene}} & Country211
& \textcolor{clipgray}{17.0} & \textcolor{clipgray}{0.0}
& 19.2 & 8.9 & 14.7 & 2.5 & 15.1 & 0.0 & 15.2 & 1.9
& 15.8 & 0.3 & 10.2 & 4.4 & 13.3 & 1.1 & 16.7 & 14.0
& \textcolor{blue!80}{-0.3} & \textcolor{red!80}{+14.0} \\
\midrule

& FGVCAircraft
& \textcolor{clipgray}{23.1} & \textcolor{clipgray}{0.0}
& 23.8 & 16.7 & 17.8 & 7.2 & 18.8 & 1.3 & 20.4 & 6.0
& 23.3 & 4.9 & 13.8 & 10.6 & 18.3 & 1.5 & 20.7 & 22.4
& \textcolor{blue!80}{-2.4} & \textcolor{red!80}{+22.4} \\
\rowcolor{gray!10}
& EuroSAT
& \textcolor{clipgray}{42.9} & \textcolor{clipgray}{0.0}
& 29.6 & 22.3 & 41.6 & 4.9 & 41.7 & 8.5 & 38.7 & 25.5
& 42.3 & 8.3 & 45.6 & 10.2 & 40.0 & 12.8 & 42.2 & 42.0
& \textcolor{blue!80}{-0.7} & \textcolor{red!80}{+42.0} \\
& DTD
& \textcolor{clipgray}{42.3} & \textcolor{clipgray}{0.1}
& 44.2 & 36.2 & 40.5 & 26.7 & 40.4 & 8.5 & 40.6 & 22.1
& 41.9 & 20.6 & 35.7 & 22.1 & 37.4 & 14.9 & 41.2 & 34.7
& \textcolor{blue!80}{-1.1} & \textcolor{red!80}{+34.6} \\
\rowcolor{gray!10}
\multirow{-4}{*}{\rotatebox[origin=c]{90}{Domain}} & PCAM
& \textcolor{clipgray}{48.4} & \textcolor{clipgray}{7.4}
& 54.6 & 39.2 & 48.6 & 48.4 & 48.4 & 36.1 & 48.5 & 48.2
& 44.3 & 4.3 & 48.2 & 23.3 & 48.9 & 48.0 & 48.6 & 52.8
& \textcolor{red!80}{+0.2} & \textcolor{red!80}{+45.4} \\
\midrule

\rowcolor{lightblue}
All & Avg.
& \textcolor{clipgray}{63.6} & \textcolor{clipgray}{1.0}
& \textbf{63.3} & 51.2 & 56.7 & 34.5 & 61.0 & 11.0 & 61.2 & 34.4
& 63.2 & 23.7 & 55.0 & 24.5 & 58.0 & 19.6 & 61.1 & \textbf{52.3}
& \textcolor{blue!80}{-2.5} & \textcolor{red!80}{+51.3} \\
\bottomrule
\end{tabular}}
\end{table*}

\textbf{Efficiency Analysis.}
Table~\ref{tab:inference_latency} compares the inference latency of test-time defenses. Benefiting from selective rectification, BaP processes clean inputs in only 3.75\,ms, while adversarial inputs require 57.37\,ms. Its average latency is 30.56\,ms, making it \(5.7\times\) faster than R-TPT and \(6\times\) faster than HD. Although TTE and ET3 are faster, their robust accuracies are only 23.7\% and 19.6\%, respectively. Table~\ref{tab:stage_latency} further reports the runtime of BaP stages. Notably, the closed-form edit implants the probe in only 2.70\,ms. Sample detection requires only 3.56\,ms per image, while the 53.94\,ms rectification is invoked only for suspicious inputs. Overall, BaP balances adversarial robustness with inference efficiency, supporting practical deployment in real-world scenarios.
 
\begin{table*}[t]
\centering
\caption{Efficiency analysis on an NVIDIA GeForce RTX 4090 GPU.}
\label{tab:efficiency}
\small

\begin{subtable}[t]{0.68\textwidth}
  \vspace{0pt}
  \centering
  \caption{End-to-end inference latency (ms/image).}
  \label{tab:inference_latency}
  \setlength{\tabcolsep}{3.5pt}
  \renewcommand{\arraystretch}{1.05}
  \resizebox{\linewidth}{!}{
    \begin{tabular}{@{}lcccccccc@{}}
      \toprule
      \textbf{Time}
      & \textcolor{clipgray}{CLIP}
      & R-TPT
      & HD
      & Anti-Adv
      & TTE
      & TTC
      & ET3
      & BaP \\
      \midrule
      \textbf{$T_{\mathit{Clean}}$}
      & \textcolor{clipgray}{3.37}
      & 176.45
      & 184.43
      & 32.39
      & 5.43
      & 33.24
      & 23.17
      & \textbf{3.75} \\

      $T_{\mathit{Rob.}}$
      & \textcolor{clipgray}{3.37}
      & 176.20
      & 189.34
      & 32.44
      & \textbf{5.41}
      & 35.13
      & 23.58
      & 57.37 \\

      \rowcolor{lightblue}
      $T_{\mathit{Avg.}}$
      & \textcolor{clipgray}{3.37}
      & 176.33
      & 186.89
      & 32.42
      & \textbf{5.42}
      & 34.19
      & 23.38
      & 30.56 \\
      \bottomrule
    \end{tabular}
  }
\end{subtable}
\hfill
\begin{subtable}[t]{0.31\textwidth}
  \vspace{0pt}
  \centering
  \caption{Runtime of BaP stages (ms).}
  \label{tab:stage_latency}
  \setlength{\tabcolsep}{3.5pt}
  \renewcommand{\arraystretch}{1.05}
  \begin{tabular}{@{}lcc@{}}
    \toprule
    \textbf{Stage} & \textbf{Time} \\
    \midrule
    Rank-one Implantation  & 2.70 \\
    Sample Detection     & 3.56 \\
    Sample Rectification & 53.94 \\
    \bottomrule
  \end{tabular}
\end{subtable}
\vspace{-0.4cm}
\end{table*}

\begin{wraptable}{r}{7cm}
\vspace{-1.5\baselineskip}
\centering
\small
\setlength{\tabcolsep}{1.3pt}
\renewcommand{\arraystretch}{1.1}
\setlength{\belowcaptionskip}{5pt}
\caption{ImageNet top-1 accuracy (\%) across attack objectives at $\ell_\infty=4/255$ with 50 steps. `AA' denotes `AutoAttack'.}
\label{tab:attack_objectives}
\resizebox{\linewidth}{!}{
    \begin{tabular}{lrrrrrrrr}
    \toprule
    \multirow{2.5}{*}{\textbf{Method}} & \multirow{2.5}{*}{\textbf{Clean}} & \multicolumn{2}{c}{\textbf{Cross-modal}} & \multicolumn{3}{c}{\textbf{Targeted}} & \multicolumn{2}{c}{\textbf{Label-free}} \\
    \cmidrule(lr){3-4}\cmidrule(lr){5-7}\cmidrule(lr){8-9}
    & & PGD & AA & PGD & DLR & AA & PGD & AA \\
    \midrule \textcolor{clipgray}{CLIP}
    & \textcolor{clipgray}{63.9}
    & \textcolor{clipgray}{0.0}
    & \textcolor{clipgray}{0.0}
    & \textcolor{clipgray}{0.0}
    & \textcolor{clipgray}{0.0}
    & \textcolor{clipgray}{0.0}
    & \textcolor{clipgray}{0.1}
    & \textcolor{clipgray}{0.0} \\
    TTC & 40.9 & 2.7 & 0.3 & 25.4 & 6.8 & 6.2 & 10.2 & 1.6 \\
    ET3 & 58.6 & 3.3 & 0.9 & 24.5 & 16.5 & 2.3 & 15.9 & 8.5 \\
    \rowcolor{lightblue}
    BaP
    & \textbf{60.3}\drop{3.6}
    & \textbf{34.6}\gain{34.6}
    & \textbf{35.1}\gain{35.1}
    & \textbf{36.1}\gain{36.1}
    & \textbf{39.3}\gain{39.3}
    & \textbf{36.9}\gain{36.9}
    & \textbf{42.4}\gain{42.3}
    & \textbf{40.0}\gain{40.0} \\

    \bottomrule
    \end{tabular}
}
\vspace{-0.5\baselineskip}
\end{wraptable}

\begin{table*}[t]
\centering
\caption{Comparison of zero-shot classification accuracy under stronger attacks at $\ell_\infty=4/255$. }
\label{tab:strong_attacks}
\setlength{\tabcolsep}{1.5pt}
\resizebox{\textwidth}{!}{%
\begin{tabular}{l|cccccccccccc}
\toprule
\multirow{2.5}{*}{\textbf{Method}} & \multicolumn{3}{c}{\textbf{General}} & \multicolumn{3}{c}{\textbf{Fine-Grained}} & \multicolumn{3}{c}{\textbf{Scene}} & \multicolumn{3}{c}{\textbf{Domain}} \\
\cmidrule(lr){2-4}\cmidrule(lr){5-7}\cmidrule(lr){8-10}\cmidrule(lr){11-13}
& Clean & PGD & AutoAttack & Clean & PGD & AutoAttack & Clean & PGD & AutoAttack & Clean & PGD & AutoAttack \\
\midrule
\textcolor{clipgray}{CLIP}
& \textcolor{clipgray}{79.4}
& \textcolor{clipgray}{0.0}
& \textcolor{clipgray}{0.0}
& \textcolor{clipgray}{76.2}
& \textcolor{clipgray}{0.0}
& \textcolor{clipgray}{0.0}
& \textcolor{clipgray}{40.3}
& \textcolor{clipgray}{0.0}
& \textcolor{clipgray}{0.0}
& \textcolor{clipgray}{39.2}
& \textcolor{clipgray}{0.0}
& \textcolor{clipgray}{0.0} \\
\rowcolor{gray!10}
R{-}TPT & 78.5 & 36.2 & 28.4 & \textbf{76.7} & \textbf{46.4} & \textbf{42.8} & \textbf{41.7} & \textbf{26.7} & \textbf{26.4} & 38.1 & 27.5 & \textbf{23.2} \\
TTE & \textbf{80.7} & 13.4 & 11.5 & 73.6 & 1.7 & 0.4 & 40.6 & 1.2 & 1.4 & 38.0 & 13.4 & 11.5 \\
\rowcolor{gray!10}
LPF & 78.6 & 37.4 & 30.8 & 68.2 & 11.3 & 8.2 & 37.0 & 8.0 & 6.9 & 37.1 & 18.6 & 16.7 \\

HD & 76.9 & 0.5 & 0.1 & 72.8 & 0.0 & 0.0 & 37.4 & 0.0 & 0.0 & 37.3 & 0.1 & 0.0 \\
\rowcolor{gray!10}
Anti{-}Adv & 76.2 & 22.6 & 2.7 & 74.1 & 1.7 & 0.0 & 38.9 & 1.1 & 0.2 & 37.1 & 9.8 & 1.7 \\
TTC & 73.3 & 14.7 & 0.6 & 59.7 & 2.0 & 0.0 & 28.9 & 1.2 & 0.0 & 35.8 & 9.3 & 0.3 \\
\rowcolor{gray!10}
ET3 & 72.9 & 7.7 & 0.3 & 69.0 & 0.9 & 0.0 & 35.1 & 0.4 & 0.0 & 36.1 & 8.7 & 0.1 \\

\rowcolor{lightblue}
BaP
& 76.9\drop{2.5}
& \textbf{57.2}\gain{57.2}
& \textbf{35.3}\gain{35.3}
& 71.6\drop{4.6}
& 43.1\gain{43.1}
& 17.4\gain{17.4}
& 38.6\drop{1.7}
& 22.4\gain{22.4}
& 8.2\gain{8.2}
& \textbf{38.2}\drop{1.0}
& \textbf{29.1}\gain{29.1}
& 13.8\gain{13.8} \\

\bottomrule
\end{tabular}}
\vspace{-0.5\baselineskip}
\end{table*}


\definecolor{bapblue}{RGB}{226,239,252}

\begin{table*}[t]
    \centering
    \small
    
    \setlength{\tabcolsep}{1.3pt}
    \renewcommand{\arraystretch}{1.15}
    \caption{Comparison of zero-shot classification accuracy on CLIP-B/32 and CLIP-L/14 under 10-step PGD at $\ell_\infty=1/255$. More detailed results are provided in Appendix~\ref{app:transfer_results}.}
    \resizebox{\textwidth}{!}{%
    \begin{tabular}{l|cccccccc|cccccccc}
        \toprule

        \multirow{4}{*}{\textbf{Method}}
        & \multicolumn{8}{c|}{\textbf{CLIP-B/32}}
        & \multicolumn{8}{c}{\textbf{CLIP-L/14}} \\
        \cmidrule(lr){2-9}
        \cmidrule(lr){10-17}

        & \multicolumn{2}{c}{\textbf{General}}
        & \multicolumn{2}{c}{\textbf{FG}}
        & \multicolumn{2}{c}{\textbf{Scene}}
        & \multicolumn{2}{c|}{\textbf{Domain}}
        & \multicolumn{2}{c}{\textbf{General}}
        & \multicolumn{2}{c}{\textbf{FG}}
        & \multicolumn{2}{c}{\textbf{Scene}}
        & \multicolumn{2}{c}{\textbf{Domain}} \\
        \cmidrule(lr){2-17}

        & clean & rob
        & clean & rob
        & clean & rob
        & clean & rob
        & clean & rob
        & clean & rob
        & clean & rob
        & clean & rob \\
        \midrule

        \textcolor{clipgray}{CLIP}
        & \textcolor{clipgray}{76.7}
        & \textcolor{clipgray}{4.0}
        & \textcolor{clipgray}{71.8}
        & \textcolor{clipgray}{0.3}
        & \textcolor{clipgray}{38.8}
        & \textcolor{clipgray}{0.4}
        & \textcolor{clipgray}{35.9}
        & \textcolor{clipgray}{6.6}
        & \textcolor{clipgray}{83.7}
        & \textcolor{clipgray}{4.0}
        & \textcolor{clipgray}{84.2}
        & \textcolor{clipgray}{0.3}
        & \textcolor{clipgray}{45.3}
        & \textcolor{clipgray}{0.2}
        & \textcolor{clipgray}{46.8}
        & \textcolor{clipgray}{0.3} \\
        \rowcolor{gray!10}
        R-TPT
        & 72.9 & 41.9
        & \textbf{71.4} & 45.4
        & 38.7 & 29.6
        & 35.7 & 27.2
        & 84.2 & \textbf{76.8}
        & \textbf{83.5} & \textbf{70.3}
        & 47.1 & 37.5
        & 42.4 & \textbf{37.3} \\

        LPF
        & 74.8 & 38.0
        & 62.0 & 18.9
        & 36.2 & 11.3
        & 33.7 & 17.4
        & 84.0 & 66.9
        & 78.8 & 51.7
        & 44.2 & 26.2
        & 44.5 & 28.5 \\
\rowcolor{gray!10}
        HD
        & 76.0 & 15.9
        & 68.6 & 4.9
        & 35.1 & 3.1
        & 35.1 & 13.6
        & 82.7 & 38.3
        & 79.7 & 9.3
        & 43.0 & 6.2
        & 44.2 & 15.9 \\

        Anti-Adv
        & 75.3 & 39.1
        & 70.3 & 12.6
        & 37.3 & 7.0
        & 33.2 & 19.9
        & 82.5 & 67.9
        & 82.4 & 48.0
        & 45.2 & 22.0
        & 43.5 & 30.9 \\
        \rowcolor{gray!10}
        TTE
        & \textbf{77.5} & 47.3
        & 68.6 & 27.9
        & \textbf{39.3} & 9.6
        & \textbf{36.8} & 19.3
        & \textbf{86.1} & 58.5
        & 81.6 & 32.3
        & \textbf{47.8} & 12.2
        & 43.7 & 23.6 \\
        
        TTC
        & 74.5 & 43.5
        & 66.5 & 27.5
        & 32.9 & 17.5
        & 34.2 & 22.4
        & 80.5 & 33.3
        & 70.3 & 39.7
        & 35.1 & 19.7
        & 42.0 & 15.5 \\
\rowcolor{gray!10}
        ET3
        & 69.4 & 34.4
        & 62.0 & 15.6
        & 35.0 & 8.8
        & 31.3 & 20.5
        & 81.1 & 43.3
        & 77.9 & 16.4
        & 43.4 & 8.8
        & 43.2 & 22.0\\

        \rowcolor{bapblue}
    \textbf{BaP}
    & 74.8\drop{1.9}
    & \textbf{62.4}\gain{58.4}
    & 66.0\drop{5.8}
    & \textbf{54.9}\gain{54.6}
    & 37.5\drop{1.3}
    & \textbf{31.0}\gain{30.6}
    & 34.8\drop{1.1}
    & \textbf{32.7}\gain{26.1}
    & 82.1\drop{1.6}
    & 62.4\gain{58.4}
    & 79.8\drop{4.4}
    & 70.1\gain{69.8}
    & 44.4\drop{0.9}
    & \textbf{40.7}\gain{40.5}
    & \textbf{44.9}\drop{1.9}
    & 35.5\gain{35.2} \\
        
        \bottomrule
    \end{tabular}%
    }
    \label{tab:backbone_transfer}
\end{table*}






\begin{table*}[t]
    \centering
    \label{tab:additional_analyses}

    \begin{minipage}[t]{0.41\textwidth}
        \centering
        \captionof{table}{Rectification-objective ablation across  the 16 datasets.}
        \label{tab:objective_ablation}

        \footnotesize
        \setlength{\tabcolsep}{7pt}
        \renewcommand{\arraystretch}{1.25}

        \resizebox{\linewidth}{!}{%
        \begin{tabular}{ccccc}
            \toprule
            $\mathcal{L}_{e}$
            & $\mathcal{L}_{u}$
            & $\mathcal{L}_{\mathrm{esc}}$
            & \textbf{Clean}
            & \textbf{Rob.} \\
            \midrule

            $\checkmark$
            &
            &
            & \textbf{62.3}
            & 21.3 \\

            $\checkmark$
            & $\checkmark$
            &
            & \textbf{62.3}
            & 22.3 \\

            \rowcolor{lightblue}
            $\checkmark$
            & $\checkmark$
            & $\checkmark$
            & 61.1
            & \textbf{52.3} \\

            \bottomrule
        \end{tabular}%
        }
        \label{tab:ablation_a}
    \end{minipage}
    \hspace{3pt}
    \begin{minipage}[t]{0.47\textwidth}
        \centering
        \captionof{table}{The adaptive PGD at
        $\ell_\infty=1/255$ with 10 steps.}
        \label{tab:adaptive_attack}

        \footnotesize
        \setlength{\tabcolsep}{7pt}
        \renewcommand{\arraystretch}{1.15}

        \resizebox{\linewidth}{!}{%
        \begin{tabular}{@{}llcccc@{}}
            \toprule
            $\quad \quad \quad \quad \beta$
            & 
            & \textbf{0}
            & \textbf{0.01}
            & \textbf{0.1}
            & \textbf{1} \\
            \midrule

            \multirow{2}{*}{\textbf{w/o BaP}}
            & \textcolor{clipgray}{Clean}
            & \textcolor{clipgray}{63.9}
            & \textcolor{clipgray}{63.9}
            & \textcolor{clipgray}{63.9}
            & \textcolor{clipgray}{63.9} \\

            & Rob.
            & 0.0
            & 0.1
            & 0.3
            & 2.6 \\

            \addlinespace[1pt]

            \multirow{2}{*}{\textbf{w/ BaP}}
            & Clean
            & 60.3
            & 60.3
            & 60.3
            & 60.3 \\

            & Rob.
            & 39.6
            & 34.0
            & \textbf{27.2}
            & 34.7 \\

            \bottomrule
        \end{tabular}%
        }
    \end{minipage}

    \vspace{-0.2cm}
\end{table*}

\textbf{Results under Different Attack Objectives.} To evaluate the performance of BaP under different attack objectives, we test it against cross-modal, targeted, and label-free attacks with a perturbation budget of \(\ell_{\infty}=4/255\). As shown in Table~\ref{tab:attack_objectives}, BaP achieves the highest robust accuracy across all seven attack configurations. Although the implanted probe is only trained on ImageNet with PGD attack of \(\ell_{\infty}=1/255\), the consistent improvements across objectives indicate that the probe captures transferable adversarial behavior rather than being tied to a specific attack. Appendix~\ref{app:diff_att_obj} shows more details.

\textbf{Results under Stronger Attacks.} Table~\ref{tab:strong_attacks} evaluates the defenses under a larger perturbation budget of \(\ell_{\infty}=4/255\) using PGD and AutoAttack. Under PGD, BaP achieves the highest accuracy on the General and Domain benchmarks, reaching 57.2\% and 29.1\%, respectively, while remaining competitive on Fine-Grained and Scene recognition. Under AutoAttack, BaP performs best on the General benchmarks, whereas R-TPT remains stronger on Fine-Grained, Scene, and Domain benchmarks. These results indicate that BaP remains effective under stronger attacks.

\textbf{Results on Different Backbones.} Table~\ref{tab:backbone_transfer} evaluates the transferability of BaP to CLIP ViT-B/32 and ViT-L/14. On ViT-B/32, BaP achieves the highest robust accuracy across all four dataset categories. When applied to ViT-L/14, BaP obtains the best robustness on Scene recognition at 40.7\% and achieves performance competitive with R-TPT. These results show that BaP remains effective across different CLIP architectures.

\begin{figure}[!t]
\centering
\includegraphics[width=\linewidth]{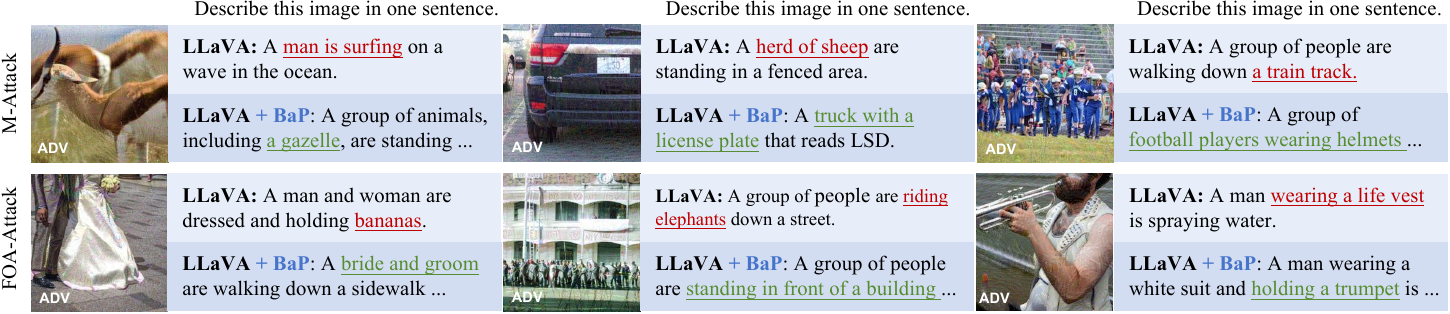}
\caption{Qualitative results of BaP against M-Attack and FOA-Attack on image captioning.}
\label{fig:LVLM_Attack}
\vspace{-0.4cm}
\end{figure}

\begin{wraptable}{r}{7cm}
\vspace{-1.5\baselineskip}
\centering
\footnotesize
\setlength{\tabcolsep}{1.7pt}
\renewcommand{\arraystretch}{0.9}
\setlength{\belowcaptionskip}{5pt}
\caption{Defense performance against M-Attack and FOA-Attack.}
\label{tab:defense_attack}
\begin{tabular}{lcccc}
\toprule
\multirow{2}{*}{\textbf{Method}}
& \multicolumn{2}{c}{\textbf{M-Attack}}
& \multicolumn{2}{c}{\textbf{FOA-Attack}} \\
\cmidrule(lr){2-3}
\cmidrule(lr){4-5}
& \textbf{Clean} & \textbf{Rob.}
& \textbf{Clean} & \textbf{Rob.} \\
\midrule

\textcolor{clipgray}{Origin Model}
& \textcolor{clipgray}{100.0} & \textcolor{clipgray}{21.1}
& \textcolor{clipgray}{100.0} & \textcolor{clipgray}{19.1} \\

+TTC defense
& 84.4 & 23.8
& 84.4 & 20.0 \\

+ET3 defense
& 87.2 & 19.4
& 87.2 & 16.7 \\

\rowcolor{lightblue}
+BaP defense
& \textbf{99.9}\drop{0.1}
& \textbf{24.6}\gain{3.5}
& \textbf{99.9}\drop{0.1}
& \textbf{23.6}\gain{4.5} \\
\bottomrule
\end{tabular}
\label{tab:lvlm_attacks}
\end{wraptable}

\textbf{Results on Attacks against LVLMs.}
Recent attacks on large vision-language models (LVLMs) have demonstrated their effectiveness against open-ended generation tasks. To evaluate BaP beyond classification, we assess the robustness of LLaVA on image captioning under M-Attack~\citep{li2025a} and FOA-Attack~\citep{jia2025adversarial}. Following their original attack settings, we randomly sample 100 images from COCO~\citep{Lin2014MicrosoftCC} and generate adversarial examples with an $\ell_\infty$ budget of $16/255$. GPTScore~\citep{li2025a} is used to evaluate captioning performance, where the detailed prompt is provided in Appendix~\ref{app:prompt}.  Table~\ref{tab:lvlm_attacks} compares BaP with TTC and ET3. Benefiting from its selective rectification mechanism, BaP preserves a clean score of 99.9, only 0.1 below the undefended model. Under M-Attack and FOA-Attack, BaP improves the robust score from 21.1 to 24.6 and from 19.1 to 23.6, respectively, outperforming both competing defenses. Fig.~\ref{fig:LVLM_Attack} further provides qualitative comparisons. Under both attacks, the undefended LLaVA produces captions containing attack-induced  concepts, whereas BaP recovers key semantics consistent with the visual content. The ROC curves for the detection are provided in Appendix~\ref{app:auc}.

\subsection{Ablation Study}

\begin{figure}[t]
\centering
\includegraphics[width=\linewidth]{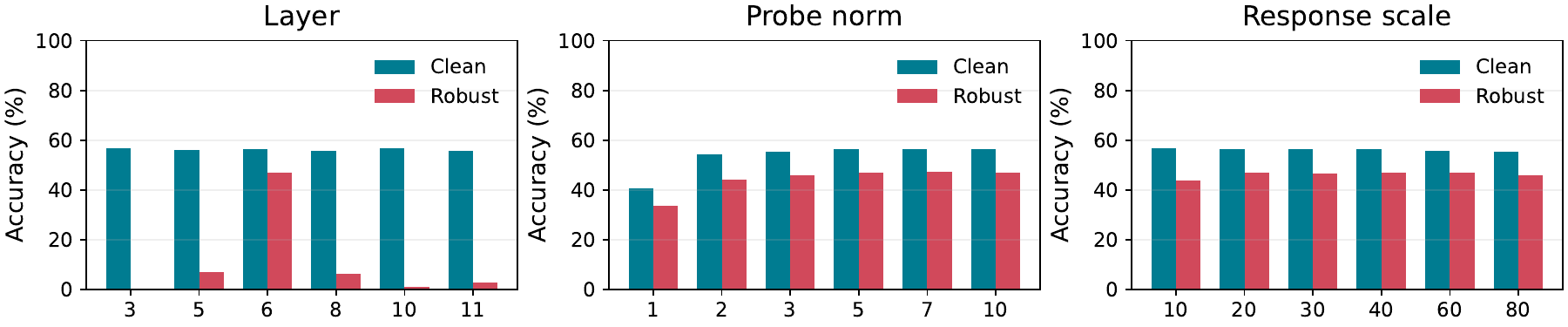}
\caption{The sensitivity to the edited layer $l$, probe norm $\nu$, and response scale $\gamma$. More ablation results are in Appendix~\ref{app:model_ablation}.}
\label{fig:hyperparameter_sensitivity}
\vspace{-0.4cm}
\end{figure}

\textbf{Ablation of the Rectification Objective.}
Table~\ref{tab:ablation_a} ablates the three objectives used for adversarial rectification. Using only the clean-subspace residual loss \(\mathcal{L}_{e}\) yields robust accuracy of 21.3\%. Adding the clean-direction alignment loss \(\mathcal{L}_{u}\) provides an improvement to 22.3\%.  Incorporating the escape loss \(\mathcal{L}_{\mathrm{esc}}\) increases robust accuracy to 52.3\%, with a limited clean accuracy reduction from 62.3\% to 61.1\%. Combining all three objectives achieves the best overall performance.

\textbf{Hyperparameter Sensitivity.}
Fig.~\ref{fig:hyperparameter_sensitivity} studies the effects of the edited layer $l$, probe norm $\nu$, and response scale $\gamma$. The edited layer has the largest influence on robustness. Placing the probe at Layer 6 yields the highest robust accuracy, whereas earlier or later layers perform considerably worse despite similar clean accuracy. Increasing the probe norm from 1 to moderate values improves both clean and robust accuracy, after which the performance gradually saturates. In contrast, BaP remains stable across response scales ranging from 10 to 80, with the best overall performance obtained around 40. Based on these results, we use Layer 6, a probe norm of 5, and a response scale of 40 as the default configuration.

\subsection{Robustness to an Adaptive Attack}
We evaluate BaP against an adaptive attack that jointly maximizes the cross-entropy loss of the edited model and suppresses the probe response. Given an adversarial input \(x_{\mathrm{adv}}\), the attack maximizes
\begin{equation}
\mathcal{L}_{\mathrm{adapt}}(x_{\mathrm{adv}},y)
=
\operatorname{CE}\!\left(f_{\theta'}(x_{\mathrm{adv}}),y\right)
-
\beta\operatorname{ReLU}\!\left(
p(x_{\mathrm{adv}})-\tau
\right),
\label{eq:adaptive_attack}
\end{equation}
where \(p(\cdot)\) denotes the probe response of Eq.~\ref{eq:p_x}, \(\tau\) is the detection threshold, and \(\beta\) controls the strength of regularization term. Table~\ref{tab:adaptive_attack} reports the results on ImageNet dataset. Increasing \(\beta\) from 0 to 0.1 reduces the robust accuracy of BaP from 39.6\% to 27.2\%, showing that explicitly targeting the detector weakens the defense. Nevertheless, at \(\beta=0.1\), BaP retains robust accuracy of 27.2\%, compared with 0.3\% without BaP. Further increasing $\beta$ to 1 raises the robust accuracy of BaP to 34.7\%. It is noted that the robust accuracy without BaP increases from 0.0\% to 2.6\%. This suggests that a larger $\beta$ weakens the attack, allowing BaP to achieve higher robust accuracy.

\section{Conclusion}

We propose BaP, a test-time defense that repurposes a controlled backdoor as an internal probe for CLIP. Through a closed-form edit, BaP treats adversarial activation shifts as triggers and maps the induced responses into a defender-specified target space. The resulting probe detects and rectifies adversarial inputs while largely preserving clean performance. Experiments across 16 benchmarks demonstrate its robustness, efficiency, and transferability across backbones and attacks. We hope BaP motivates further exploration of backdoor-inspired probes for improving adversarial robustness.

\subsection*{Ethics Statement}
Although BaP is inspired by backdoor mechanisms, it is intended for adversarial defense. Its defender-controlled probe is used to detect and rectify adversarial inputs. It should be applied only to models for which the defender has authorization. We believe this work advances the defensive use of backdoor-inspired mechanisms and contributes to more robust vision-language models.

\bibliography{iclr2027_conference}

@inproceedings{CLIP,
  title={Learning Transferable Visual Models From Natural Language Supervision},
  author={Alec Radford and Jong Wook Kim and Chris Hallacy and Aditya Ramesh and Gabriel Goh and Sandhini Agarwal and Girish Sastry and Amanda Askell and Pamela Mishkin and Jack Clark and Gretchen Krueger and Ilya Sutskever},
  booktitle={International Conference on Machine Learning (ICML)},
  year={2021},
}

@inproceedings{li2023blip2,
author = {Li, Junnan and Li, Dongxu and Savarese, Silvio and Hoi, Steven},
title = {BLIP-2: bootstrapping language-image pre-training with frozen image encoders and large language models},
year = {2023},
booktitle = {Proceedings of the 40th International Conference on Machine Learning (ICML)},
articleno = {814},
numpages = {13},
location = {, Honolulu, Hawaii, USA, },
}

@article{Qwen-VL,
  title={Qwen-VL: A Versatile Vision-Language Model for Understanding, Localization, Text Reading, and Beyond},
  author={Bai, Jinze and Bai, Shuai and Yang, Shusheng and Wang, Shijie and Tan, Sinan and Wang, Peng and Lin, Junyang and Zhou, Chang and Zhou, Jingren},
  journal={arXiv preprint arXiv:2308.12966},
  year={2023}
}

@inproceedings{cui2024robustness,
  title={On the robustness of large multimodal models against image adversarial attacks},
  author={Cui, Xuanming and Aparcedo, Alejandro and Jang, Young Kyun and Lim, Ser-Nam},
  booktitle={2024 IEEE/CVF Conference on Computer Vision and Pattern Recognition (CVPR)},
  pages={24625--24634},
  year={2024},
  organization={IEEE}
}

@InProceedings{Nie_2026_CVPR,
    author    = {Nie, Sen and Zhang, Jie and Yan, Jianxin and Shan, Shiguang and Chen, Xilin},
    title     = {V-Attack: Targeting Disentangled Value Features for Controllable Adversarial Attacks on LVLMs},
    booktitle = {Proceedings of the IEEE/CVF Conference on Computer Vision and Pattern Recognition (CVPR)},
    month     = {June},
    year      = {2026},
    pages     = {42257-42267}
}

@inproceedings{
mao2023understanding,
title={Understanding Zero-shot Adversarial Robustness for Large-Scale Models},
author={Chengzhi Mao and Scott Geng and Junfeng Yang and Xin Wang and Carl Vondrick},
booktitle={The International Conference on Learning Representations (ICLR)},
year={2023},
}

@inproceedings{wang2024pre,
  title={Pre-trained model guided fine-tuning for zero-shot adversarial robustness},
  author={Wang, Sibo and Zhang, Jie and Yuan, Zheng and Shan, Shiguang},
  booktitle={Proceedings of the IEEE conference on Computer Vision and Pattern Recognition (CVPR)},
  pages={24502--24511},
  year={2024},
  organization={IEEE}
}

@inproceedings{perez2021enhancing,
  title={Enhancing adversarial robustness via test-time transformation ensembling},
  author={P{\'e}rez, Juan C and Alfarra, Motasem and Jeanneret, Guillaume and Rueda, Laura and Thabet, Ali and Ghanem, Bernard and Arbel{\'a}ez, Pablo},
  booktitle={2021 IEEE/CVF International Conference on Computer Vision Workshops (ICCVW)},
  pages={81--91},
  year={2021},
  organization={IEEE}
}

@ARTICLE{zhu2026aisheild,
  author={Zhu, Hong and Zhang, Shengzhi and Chen, Kai},
  journal={IEEE Transactions on Dependable and Secure Computing (TDSC)}, 
  title={AI-Shielder: Exploiting Backdoors to Defend Against Adversarial Attacks}, 
  year={2026},
  volume={23},
  number={1},
  pages={1244-1259}}

@inproceedings{deng2009imagenet,
  title={Imagenet: A large-scale hierarchical image database},
  author={Deng, Jia and Dong, Wei and Socher, Richard and Li, Li-Jia and Li, Kai and Fei-Fei, Li},
  booktitle={Proceedings of the IEEE conference on Computer Vision and Pattern Recognition (CVPR)},
  pages={248--255},
  year={2009},
  organization={IEEE}
}

@INPROCEEDINGS{BadVision,
  author={Liu, Zhaoyi and Zhang, Huan},
  booktitle={Proceedings of the IEEE/CVF Conference on Computer Vision and Pattern Recognition (CVPR)}, 
  title={Stealthy Backdoor Attack in Self-Supervised Learning Vision Encoders for Large Vision Language Models}, 
  year={2025},
  volume={},
  number={},
  doi={10.1109/CVPR52734.2025.02333}}

@InProceedings{He_2016_CVPR,
author = {He, Kaiming and Zhang, Xiangyu and Ren, Shaoqing and Sun, Jian},
title = {Deep Residual Learning for Image Recognition},
booktitle = {Proceedings of the IEEE conference on Computer Vision and Pattern Recognition (CVPR)},
month = {June},
year = {2016}
}

@inproceedings{shan2020gotta,
  title={Gotta catch'em all: Using honeypots to catch adversarial attacks on neural networks},
  author={Shan, Shawn and Wenger, Emily and Wang, Bolun and Li, Bo and Zheng, Haitao and Zhao, Ben Y},
  booktitle={Proceedings of the 2020 ACM SIGSAC conference on computer and communications security},
  pages={67--83},
  year={2020}
}

@inproceedings{gu2017identifying,
  title={Identifying vulnerabilities in the machine learning model supply chain},
  author={Tianyu Gu and Brendan Dolan-Gavitt and Siddharth Garg},
  booktitle={Proceedings of the Neural Information Processing Symposium Workshop Mach. Learning Security (MLSec)},
  pages={1--5},
  year={2017}
}

@inproceedings{xing2025clip,
  title={Clip is strong enough to fight back: Test-time counterattacks towards zero-shot adversarial robustness of clip},
  author={Xing, Songlong and Zhao, Zhengyu and Sebe, Nicu},
  booktitle={Proceedings of the IEEE conference on Computer Vision and Pattern Recognition (CVPR)},
  pages={15172--15182},
  year={2025},
  organization={IEEE}
}

@article{krizhevsky2009learning,
  title={Learning multiple layers of features from tiny images},
  author={Krizhevsky, Alex},
  journal={Technical Report, University of Toronto},
  year={2009},
  publisher={Toronto, ON, Canada}
}

@inproceedings{coates2011analysis,
  title={An analysis of single-layer networks in unsupervised feature learning},
  author={Coates, Adam and Ng, Andrew and Lee, Honglak},
  booktitle={Proceedings of the fourteenth international conference on artificial intelligence and statistics},
  pages={215--223},
  year={2011},
  organization={JMLR Workshop and Conference Proceedings}
}

@inproceedings{fei2004learning,
  title={Learning generative visual models from few training examples: An incremental bayesian approach tested on 101 object categories},
  author={Fei-Fei, Li and Fergus, Rob and Perona, Pietro},
  booktitle={Proceedings of the IEEE conference on Computer Vision and Pattern Recognition Workshop (CVPRW)},
  pages={178--178},
  year={2004},
  organization={IEEE}
}

@techreport{griffin2007caltech,
  title={Caltech-256 object category dataset},
  author={Griffin, Gregory and Holub, Alex and Perona, Pietro and others},
  year={2007},
  institution={Technical Report 7694, California Institute of Technology Pasadena}
}

@inproceedings{parkhi2012cats,
  title={Cats and dogs},
  author={Parkhi, Omkar M and Vedaldi, Andrea and Zisserman, Andrew and Jawahar, CV},
  booktitle={Proceedings of the IEEE conference on Computer Vision and Pattern Recognition (CVPR)},
  pages={3498--3505},
  year={2012},
  organization={IEEE}
}

@inproceedings{nilsback2008automated,
  title={Automated flower classification over a large number of classes},
  author={Nilsback, Maria-Elena and Zisserman, Andrew},
  booktitle={Sixth Indian conference on computer vision, graphics \& image processing},
  pages={722--729},
  year={2008},
  organization={IEEE}
}

@inproceedings{bossard2014food,
  title={Food-101--mining discriminative components with random forests},
  author={Bossard, Lukas and Guillaumin, Matthieu and Van Gool, Luc},
  booktitle={European Conference on Computer Vision (ECCV)},
  pages={446--461},
  year={2014},
  organization={Springer}
}

@inproceedings{krause20133d,
  title={3d object representations for fine-grained categorization},
  author={Krause, Jonathan and Stark, Michael and Deng, Jia and Fei-Fei, Li},
  booktitle={Proceedings of the IEEE International Conference on Computer Vision Workshops (ICCVW)},
  pages={554--561},
  year={2013}
}

@inproceedings{xiao2010sun,
  title={Sun database: Large-scale scene recognition from abbey to zoo},
  author={Xiao, Jianxiong and Hays, James and Ehinger, Krista A and Oliva, Aude and Torralba, Antonio},
  booktitle={Proceedings of the IEEE conference on Computer Vision and Pattern Recognition (CVPR)},
  pages={3485--3492},
  year={2010},
  organization={IEEE}
}

@article{maji2013fine,
  title={Fine-grained visual classification of aircraft},
  author={Maji, Subhransu and Rahtu, Esa and Kannala, Juho and Blaschko, Matthew and Vedaldi, Andrea},
  journal={arXiv preprint arXiv:1306.5151},
  year={2013}
}

@article{helber2019eurosat,
  title={Eurosat: A novel dataset and deep learning benchmark for land use and land cover classification},
  author={Helber, Patrick and Bischke, Benjamin and Dengel, Andreas and Borth, Damian},
  journal={IEEE Journal of Selected Topics in Applied Earth Observations and Remote Sensing},
  volume={12},
  number={7},
  pages={2217--2226},
  year={2019},
  publisher={IEEE}
}

@inproceedings{cimpoi2014describing,
  title={Describing textures in the wild},
  author={Cimpoi, Mircea and Maji, Subhransu and Kokkinos, Iasonas and Mohamed, Sammy and Vedaldi, Andrea},
  booktitle={Proceedings of the IEEE conference on Computer Vision and Pattern Recognition (CVPR)},
  pages={3606--3613},
  year={2014}
}

@inproceedings{veeling2018rotation,
  title={Rotation equivariant CNNs for digital pathology},
  author={Veeling, Bastiaan S and Linmans, Jasper and Winkens, Jim and Cohen, Taco and Welling, Max},
  booktitle={International Conference on Medical image computing and computer-assisted intervention (MICCAI)},
  pages={210--218},
  year={2018},
  organization={Springer}
}

@inproceedings{wang2025tapt,
  title={Tapt: Test-time adversarial prompt tuning for robust inference in vision-language models},
  author={Wang, Xin and Chen, Kai and Zhang, Jiaming and Chen, Jingjing and Ma, Xingjun},
  booktitle={Proceedings of the IEEE conference on Computer Vision and Pattern Recognition (CVPR)},
  pages={19910--19920},
  year={2025}
}

@inproceedings{sheng2025r,
  title={R-TPT: Improving Adversarial Robustness of Vision-Language Models through Test-Time Prompt Tuning},
  author={Sheng, Lijun and Liang, Jian and Wang, Zilei and He, Ran},
  booktitle={Proceedings of the IEEE conference on Computer Vision and Pattern Recognition (CVPR)},
  pages={29958--29967},
  year={2025}
}

@article{schlarmann2024robustclip,
    title={Robust CLIP: Unsupervised Adversarial Fine-Tuning of Vision Embeddings for Robust Large Vision-Language Models}, 
    author={Christian Schlarmann and Naman Deep Singh and Francesco Croce and Matthias Hein},
    year={2024},
    journal={ICML}
}

@inproceedings{alfarra2022combating,
  title={Combating adversaries with anti-adversaries},
  author={Alfarra, Motasem and P{\'e}rez, Juan C and Thabet, Ali and Bibi, Adel and Torr, Philip HS and Ghanem, Bernard},
  booktitle={Proceedings of the AAAI Conference on Artificial Intelligence (AAAI)},
  volume={36},
  pages={5992--6000},
  year={2022}
}

@inproceedings{wanglearning,
  title={Learning Robust Vision-Language Models from Natural Latent Spaces},
  author={Wang, Zhangyun and Ding, Ni and Mahanti, Aniket},
  booktitle={Advances in Neural Information Processing Systems (NeurIPS)},
  year={2025}
}

@article{zhou2024few,
  title={Few-shot adversarial prompt learning on vision-language models},
  author={Zhou, Yiwei and Xia, Xiaobo and Lin, Zhiwei and Han, Bo and Liu, Tongliang},
  journal={Advances in Neural Information Processing Systems (NeurIPS)},
  volume={37},
  pages={3122--3156},
  year={2024}
}

@inproceedings{zhang2024adversarial,
  title={Adversarial prompt tuning for vision-language models},
  author={Zhang, Jiaming and Ma, Xingjun and Wang, Xin and Qiu, Lingyu and Wang, Jiaqi and Jiang, Yu-Gang and Sang, Jitao},
  booktitle={European conference on computer vision (ECCV)},
  pages={56--72},
  year={2024},
  organization={Springer}
}

@inproceedings{li2024one,
  title={One prompt word is enough to boost adversarial robustness for pre-trained vision-language models},
  author={Li, Lin and Guan, Haoyan and Qiu, Jianing and Spratling, Michael},
  booktitle={Proceedings of the IEEE conference on Computer Vision and Pattern Recognition (CVPR)},
  pages={24408--24419},
  year={2024}
}

@article{zhu2025enhancing,
  title={Enhancing CLIP Robustness via Cross-Modality Alignment},
  author={Zhu, Xingyu and Zhu, Beier and Wang, Shuo and Zhao, Kesen and Zhang, Hanwang},
  journal={The Annual Conference on Neural Information Processing Systems (NeurIPS)},
  year={2025}
}

@inproceedings{DBLP:conf/iclr/ZhangB0GC25,
  author       = {Mingkun Zhang and
                  Keping Bi and
                  Wei Chen and
                  Jiafeng Guo and
                  Xueqi Cheng},
  title        = {CLIPure: Purification in Latent Space via {CLIP} for Adversarially
                  Robust Zero-Shot Classification},
  booktitle    = {The International Conference on Learning Representations (ICLR)},
  year         = {2025}
}

@article{wu2021attacking,
  title={Attacking adversarial attacks as a defense},
  author={Wu, Boxi and Pan, Heng and Shen, Li and Gu, Jindong and Zhao, Shuai and Li, Zhifeng and Cai, Deng and He, Xiaofei and Liu, Wei},
  journal={arXiv preprint arXiv:2106.04938},
  year={2021}
}

@article{ziyadinov2023low,
  title={Low-pass image filtering to achieve adversarial robustness},
  author={Ziyadinov, Vadim and Tereshonok, Maxim},
  journal={Sensors},
  volume={23},
  number={22},
  pages={9032},
  year={2023},
  publisher={MDPI}
}

@InProceedings{Mirza_2026_CVPR,
    author    = {Mirza, Mujtaba Hussain and D'Orazio, Antonio and Melamed, Odelia and Masi, Iacopo},
    title     = {A Provable Energy-Guided Test-Time Defense Boosting Adversarial Robustness of Large Vision-Language Models},
    booktitle = {Proceedings of the IEEE/CVF Conference on Computer Vision and Pattern Recognition (CVPR)},
    month     = {June},
    year      = {2026},
    pages     = {8598-8609}
}

@inproceedings{
nie2026contrastive,
title={Contrastive Spectral Rectification: Test-Time Defense towards Zero-shot Adversarial Robustness of {CLIP}},
author={Sen Nie and Jie Zhang and Zhuo Wang and Shiguang Shan and Xilin Chen},
booktitle={International Conference on Machine Learning (ICML)},
year={2026},
}

@article{shu2022test,
  title={Test-time prompt tuning for zero-shot generalization in vision-language models},
  author={Shu, Manli and Nie, Weili and Huang, De-An and Yu, Zhiding and Goldstein, Tom and Anandkumar, Anima and Xiao, Chaowei},
  journal={Advances in Neural Information Processing Systems (NeurIPS)},
  volume={35},
  pages={14274--14289},
  year={2022}
}

@article{abdul2023align,
  title={Align Your Prompts: Test-Time Prompting with Distribution Alignment for Zero-Shot Generalization},
  author={Abdul Samadh, Jameel and Gani, Mohammad Hanan and Hussein, Noor and Khattak, Muhammad Uzair and Naseer, Muhammad Muzammal and Shahbaz Khan, Fahad and Khan, Salman H.},
  journal={Advances in Neural Information Processing Systems (NeurIPS)},
  volume={36},
  pages={80396--80413},
  year={2023}
}

@inproceedings{hossain2024securing,
  title={Securing vision-language models with a robust encoder against jailbreak and adversarial attacks},
  author={Hossain, Md Zarif and Imteaj, Ahmed},
  booktitle={2024 IEEE International Conference on Big Data (BigData)},
  pages={6250--6259},
  year={2024},
  organization={IEEE}
}

@article{han2025d,
  title={D-TPT: Dimensional Entropy Maximization for Calibrating Test-Time Prompt Tuning in Vision-Language Models},
  author={Han, Jisu and Hwang, Wonjun},
  journal={arXiv preprint arXiv:2510.09473},
  year={2025}
}

@inproceedings{yoon2024c,
  title={C-tpt: Calibrated test-time prompt tuning for vision-language models via text feature dispersion},
  author={Yoon, Hee Suk and Yoon, Eunseop and Tee, Joshua Tian Jin and Hasegawa-Johnson, Mark and Li, Yingzhen and Yoo, Chang},
  booktitle={International Conference on Learning Representations (ICLR)},
  year={2024}
}

@inproceedings{madry2017towards,
  author       = {Aleksander Madry and
                  Aleksandar Makelov and
                  Ludwig Schmidt and
                  Dimitris Tsipras and
                  Adrian Vladu},
  title        = {Towards Deep Learning Models Resistant to Adversarial Attacks},
  booktitle    = {International Conference on Learning Representations (ICLR)},
  year         = {2018}
}

@inproceedings{croce2020reliable,
  title={Reliable evaluation of adversarial robustness with an ensemble of diverse parameter-free attacks},
  author={Croce, Francesco and Hein, Matthias},
  booktitle={International conference on machine learning},
  pages={2206--2216},
  year={2020},
  organization={PMLR}
}

@inproceedings{li2024badedit,
  title={Badedit: Backdooring large language models by model editing},
  author={Li, Yanzhou and Li, Tianlin and Chen, Kangjie and Zhang, Jian and Liu, Shangqing and Wang, Wenhan and Zhang, Tianwei and Liu, Yang},
  booktitle={International Conference on Learning Representations (ICLR)},
  year={2024}
}

@inproceedings{wang2024eviledit,
  title={Eviledit: Backdooring text-to-image diffusion models in one second},
  author={Wang, Hao and Guo, Shangwei and He, Jialing and Chen, Kangjie and Zhang, Shudong and Zhang, Tianwei and Xiang, Tao},
  booktitle={Proceedings of the 32nd ACM International Conference on Multimedia},
  pages={3657--3665},
  year={2024}
}

@article{meng2022locating,
  title={Locating and editing factual associations in gpt},
  author={Meng, Kevin and Bau, David and Andonian, Alex and Belinkov, Yonatan},
  journal={Advances in neural information processing systems (NeurIPS)},
  volume={35},
  pages={17359--17372},
  year={2022}
}

@inproceedings{mela-etal-2024-mass,
    title = "Mass-Editing Memory with Attention in Transformers: A cross-lingual exploration of knowledge",
    author = "Tamayo, Daniel  and
      Gonzalez-Agirre, Aitor  and
      Hernando, Javier  and
      Villegas, Marta",
    editor = "Ku, Lun-Wei  and
      Martins, Andre  and
      Srikumar, Vivek",
    booktitle = "Findings of the Association for Computational Linguistics (ACL)",
    month = aug,
    year = "2024",
    address = "Bangkok, Thailand",
    publisher = "Association for Computational Linguistics",
    pages = "5831--5847",
}

@inproceedings{
liu2026adversarial,
title={Adversarial Attacks Already Tell the Answer: Directional Bias-Guided Test-time Defense for Vision-Language Models},
author={Liangsheng Liu and Si Chen and Jiamin Wu and Weiwei Feng and Zhixin Cheng and Xiaotian Yin and Wenfei Yang and Tianzhu Zhang},
booktitle={The International Conference on Learning Representations (ICLR)},
year={2026},
}

@article{penrose1955generalized,
  title   = {A Generalized Inverse for Matrices},
  author  = {Penrose, R.},
  journal = {Mathematical Proceedings of the Cambridge Philosophical Society},
  volume  = {51},
  number  = {3},
  pages   = {406--413},
  year    = {1955},
}

@article{liu2023visual_llava,
  title={Visual instruction tuning},
  author={Liu, Haotian and Li, Chunyuan and Wu, Qingyang and Lee, Yong Jae},
  journal={arXiv preprint arXiv:2304.08485},
  year={2023}
}

@inproceedings{
    jia2025adversarial,
    title={Adversarial Attacks against Closed-Source {MLLM}s via Feature Optimal Alignment},
    author={Xiaojun Jia and Sensen Gao and Simeng Qin and Tianyu Pang and Chao Du and Yihao Huang and Xinfeng Li and Yiming Li and Bo Li and Yang Liu},
    booktitle={The Annual Conference on Neural Information Processing Systems (NeurIPS)},
    year={2025},
}

@inproceedings{
    li2025a,
    title={A Frustratingly Simple Yet Highly Effective Attack Baseline: Over 90\% Success Rate Against the Strong Black-box Models of {GPT}-4.5/4o/o1},
    author={Zhaoyi Li and Xiaohan Zhao and Dong-Dong Wu and Jiacheng Cui and Zhiqiang Shen},
    booktitle={The Annual Conference on Neural Information Processing Systems (NeurIPS)},
    year={2025},
}

@inproceedings{Lin2014MicrosoftCC,
  title={Microsoft COCO: Common Objects in Context},
  author={Tsung-Yi Lin and Michael Maire and Serge J. Belongie and James Hays and Pietro Perona and Deva Ramanan and Piotr Doll{\'a}r and C. Lawrence Zitnick},
  booktitle={Proceedings of the European Conference on Computer Vision (ECCV)},
  year={2014}}
\bibliographystyle{iclr2027_conference}

\newpage

\appendix

\section*{Appendix: Table of Contents}
\begin{description}[style=multiline, leftmargin=1.7cm, font=\bfseries, itemsep=-0.2em]
    \item[Section \ref{app:algorithm}] Algorithm of BaP \dotfill \pageref{app:algorithm}
    \item[Section \ref{app:rank_one-edit}] Details of the Weight Edit \dotfill \pageref{app:rank_one-edit}
    \item[Section \ref{app:diff_att_obj}] Details of Different Attack Objectives \dotfill \pageref{app:diff_att_obj}
    \item[Section \ref{app:model_ablation}] More Ablation Studies \dotfill \pageref{app:model_ablation}
    \item[] Detection Threshold $\tau$ \dotfill \pageref{app:tau}
    \item[] Target Prompt $p$ \dotfill \pageref{app:target}
    \item[Section \ref{app:prompt}] Detailed Prompt for GPTScore \dotfill 
    \pageref{app:prompt}
    \item[Section \ref{app:transfer_results}] Detailed Results on CLIP-B/32 and CLIP-L/14 \dotfill \pageref{app:transfer_results}
    \item[Section \ref{app:auc}] Detection ROC of BaP \dotfill \pageref{app:auc}

\end{description}

\section{Algorithm of BaP}
\label{app:algorithm}
The full algorithm of BaP is provided in Algorithm \ref{alg:bap}.

\begin{algorithm}[H]
\caption{BaP Probe Implantation, Detection, and Rectification}
\label{alg:bap}
\begin{algorithmic}[1]
\Require Image encoder $f_{\theta}$, text encoder $g_\phi$, layer weight $W_l$, clean calibration set $\mathcal X$, paired adversarial samples $\{x_i,x_i^{\mathrm{adv}}\}_{i=1}^N$ on ImageNet, target prompt $p$, and parameters $K^{in}$, $K^{out}$, $\nu$, $\gamma$, $\epsilon_c$, $\alpha$, $\lambda$, and $\tau$
\Require Test input $x$
\Ensure Edited weight $W_l'$, detection decision $G(x)$, and output image $\tilde{x}$
\State Construct the input and output activation matrices $H_l$ and $Z_l$ from $\mathcal X$
\State Compute the SVDs of $H_l$ and $Z_l$
\State Construct the low-energy subspaces $\mathcal S_l^{\mathrm{in}}$ and $\mathcal S_l^{\mathrm{out}}$
\State Compute the mean attack activation shift $\bar{\Delta}_l$
\State Compute $a_j$ for $j\in\mathcal I_l^{\mathrm{in}}$ and obtain the input-side probe direction $t_l$
\State Encode $p$ as $q$ and compute $y_0=A^\dagger q$
\State Project $y_0$ onto $\mathcal S_l^{\mathrm{out}}$ and normalize it to obtain $y_l$
\State Implant the probe using $W_l'\gets W_l+y_lt_l^\top/\|t_l\|_2^2$
\State Construct $\mathcal M_l^{Clean}$ and the global clean direction $v_l^{\mathrm{clean}}$ from $\mathcal X$
\State Compute $p(x)=\hat y_l^\top W_l'h_l(x)$
\State $G(x)\gets\mathbb I[p(x)\geq\tau]$
\If{$G(x)=1$}
    \State Sample $\xi_0$ uniformly from $[-\epsilon_c,\epsilon_c]$
    \For{$r=0,1$}
        \State Update $\xi_{r+1}$ by projected gradient ascent on $\mathcal L_{\mathrm{esc}}$
    \EndFor
    \State Compute $e_l(x+\xi_2)$ and $\mathcal L_{\mathrm{rep}}$
    \State Update $\xi_3$ by one projected gradient descent step on $\mathcal L_{\mathrm{rep}}$
    \State $\tilde{x}\gets x+\xi_3$
\Else
    \State $\tilde{x}\gets x$
\EndIf
\State \Return $W_l'$, $G(x)$, and $\tilde{x}$
\end{algorithmic}
\end{algorithm}

\section{Details of the Weight Edit}
\label{app:rank_one-edit}

This section gives the derivation and properties of the weight edit used to implant the defender-controlled probe.  Following ~\citep{meng2022locating,li2024badedit}, we consider the \texttt{fc2} map at vision-encoder layer $l$,
\begin{equation}
    z_l(x)=W_l h_l(x),
    \qquad
    W_l\in\mathbb{R}^{d_o\times d_i},
\end{equation}
where $h_l(x)\in\mathbb{R}^{d_i}$ is the input activation of the \texttt{CLS} token. Given a nonzero input-side probe direction $t_l\in\mathbb{R}^{d_i}$ and a target shift $y\in\mathbb{R}^{d_o}$, we seek a weight update that maps $t_l$ exactly to $y$ while changing the original weight as little as possible. This yields the constrained problem
\begin{equation}
    \min_{\Delta W_l}\ \|\Delta W_l\|_F^2
    \quad \text{subject to} \quad
    \Delta W_l t_l=y.
    \label{eq:rank-one-objective}
\end{equation}

\textbf{Closed-form solution.}
Introduce a Lagrange multiplier $\lambda\in\mathbb{R}^{d_o}$ and write
\begin{equation}
    \mathcal{J}(\Delta W_l,\lambda)
    =\frac{1}{2}\|\Delta W_l\|_F^2
    +\lambda^\top(y-\Delta W_l t_l).
\end{equation}
Stationarity with respect to $\Delta W_l$ gives $\Delta W_l=\lambda t_l^\top$. Substitution into the constraint produces $\lambda=y/\|t_l\|_2^2$, and hence
\begin{equation}
    \Delta W_l^\star
    =\frac{y t_l^\top}{\|t_l\|_2^2},
    \qquad
    W_l'=W_l+\Delta W_l^\star.
    \label{eq:rank-one-solution}
\end{equation}
Because the objective in Eq.~\ref{eq:rank-one-objective} is strictly convex and the constraint is affine, Eq.~\ref{eq:rank-one-solution} is the unique minimum-norm solution. It also satisfies the desired mapping exactly:
\begin{equation}
    \Delta W_l^\star t_l
    =\frac{y t_l^\top t_l}{\|t_l\|_2^2}
    =y.
\end{equation}

\textbf{Response on an arbitrary activation.}
For any activation $h\in\mathbb{R}^{d_i}$, the edit introduces the output change
\begin{equation}
\Delta z_l(h)
=
\Delta W_l h
=
y_l\frac{t_l^\top h}{\|t_l\|_2^2}
=
\frac{y_l}{\|t_l\|_2}\hat t_l^\top h,
\qquad
\hat t_l=\frac{t_l}{\|t_l\|_2}.
\label{eq:rank-one-response}
\end{equation}
Let $\hat y_l=y_l/\|y_l\|_2$. The signed response of the edited weight along the target direction is
\begin{equation}
\begin{aligned}
\rho_l(h)
&=
\hat y_l^\top W_l'h \\
&=
\hat y_l^\top W_lh
+
\frac{\|y_l\|_2}{\|t_l\|_2}
\hat t_l^\top h.
\end{aligned}
\label{eq:response-magnitude}
\end{equation}
The first term in
Eq.~\ref{eq:response-magnitude} is the pre-existing response of the
original weight, whereas the second term is the response introduced
by the edit. The output-side low-energy construction suppresses the
former on clean inputs, while the attack-aligned input direction
strengthens the latter on adversarial inputs.

\section{Details of Different Attack Objectives}
\label{app:diff_att_obj}

To further assess the versatility of BaP, we evaluate it against cross-modal, targeted and label-free attacks. Here, we adopt a threat model with a perturbation budget of $\epsilon_{\text{adv}} = 4/255$ and set the number of attack iterations to $T = 50$. All adversarial examples are generated against the editing model. The formulation of each adversarial objective is described in detail below.

\textbf{Cross-modal Attacks} aim to disrupt the alignment between visual and textual representations. Given the ground-truth class $y$ and its textual prompt $t_y$, the attack minimizes the cosine similarity between the adversarial visual feature and the corresponding text feature:
\begin{equation}
\max_{\|\delta\|_{\infty}\leq\epsilon_{\mathrm{adv}}}
-
\left\langle
\frac{f_\theta(x+\delta)}
{\|f_\theta(x+\delta)\|_2},
\frac{g_\phi(t_y)}
{\|g_\phi(t_y)\|_2}
\right\rangle,
\label{eq:cross_modal_attack}
\end{equation}
where $f_\theta(\cdot)$ and $g_\phi(\cdot)$ denote the vision encoder and text encoder, respectively. This objective pushes the adversarial visual representation away from its corresponding textual representation. We optimize it using both PGD and AutoAttack.

\textbf{Targeted Attacks} aim to force the model to classify an adversarial image as a predefined target class $y_{\mathrm{tar}}$. For PGD and AutoAttack, we maximize the negative cross-entropy loss with respect to the target class:
\begin{equation}
\max_{\|\delta\|_{\infty}\leq\epsilon_{\mathrm{adv}}}
-\mathcal{L}_{\mathrm{CE}}
\left(
\left\{
\left\langle
\frac{f_\theta(x+\delta)}
{\|f_\theta(x+\delta)\|_2},
\frac{g_\phi(t_c)}
{\|g_\phi(t_c)\|_2}
\right\rangle
\right\}_{c=1}^{C},
y_{\mathrm{tar}}
\right).
\label{eq:targeted_attack}
\end{equation}
We also consider the targeted Difference of Logits Ratio (DLR) loss:
\begin{equation}
\mathcal{L}_{\mathrm{DLR}}^{\mathrm{tar}}
=
-
\frac{s_y-s_{y_{\mathrm{tar}}}}
{s_{\pi_1}-\frac{1}{2}\left(s_{\pi_3}+s_{\pi_4}\right)},
\label{eq:targeted_dlr}
\end{equation}
where
\begin{equation}
s_i=
\left\langle
\frac{f_\theta(x+\delta)}
{\|f_\theta(x+\delta)\|_2},
\frac{g_\phi(t_i)}
{\|g_\phi(t_i)\|_2}
\right\rangle
\end{equation}
denotes the cosine similarity for class $i$, and $\pi$ sorts these similarities in descending order. Maximizing $\mathcal{L}_{\mathrm{DLR}}^{\mathrm{tar}}$ reduces the margin between the ground-truth class and the target class.

\textbf{Label-free Attacks} aim to disrupt the consistency between clean and adversarial visual representations without using ground-truth labels or textual prompts. Given a clean image $x$, the attack maximizes the negative cosine similarity between its clean and adversarial visual features:
\begin{equation}
\max_{\|\delta\|_{\infty}\leq\epsilon_{\mathrm{adv}}}
-
\left\langle
\frac{f_\theta(x+\delta)}
{\|f_\theta(x+\delta)\|_2},
\frac{f_\theta(x)}
{\|f_\theta(x)\|_2}
\right\rangle .
\label{eq:label_free_attack}
\end{equation}
This objective pushes the adversarial visual representation away from its original representation without relying on label or text information. We optimize it using both PGD and APGD.

\section{More Ablation Studies}
\label{app:model_ablation}

\subsection{Ablation Study on Detection Threshold $\tau$}
\label{app:tau}

\begin{figure}[ht]
\centering
\includegraphics[width=307.5pt]{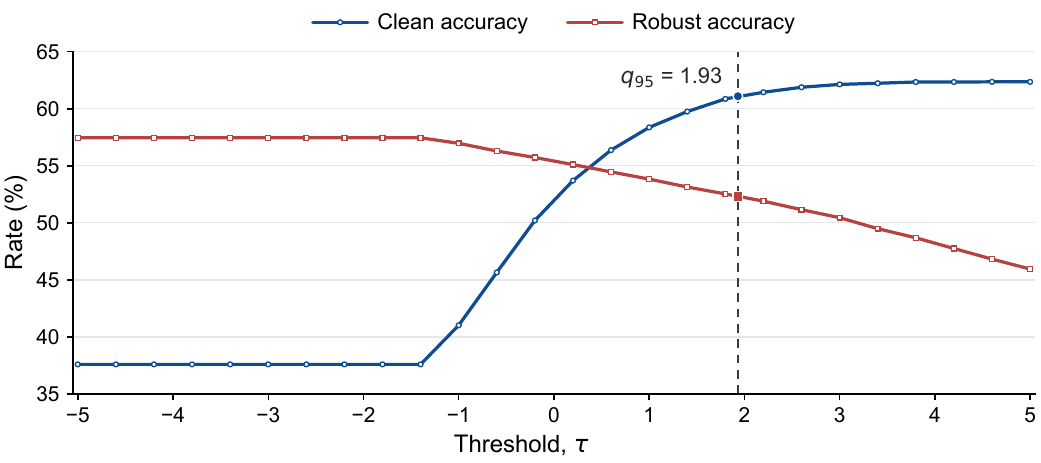}
\caption{The sensitivity to the threshold $\tau$.}
\label{fig:threshold_ablation}
\end{figure}

The detection threshold $\tau$ controls which inputs undergo rectification. Since BaP identifies an input as suspicious when $p(x)\geq\tau$, increasing $\tau$ reduces the number of rectified inputs. Fig.~\ref{fig:threshold_ablation} shows the evaluation results from -5 to 5 in increments of 0.4, where it exhibits a clear trade-off. Clean accuracy increases with $\tau$, while robust accuracy gradually decreases. The detection threshold $\tau$ is set to the 95th percentile of clean calibration scores on the held-out ImageNet calibration set, yielding $q_{95}=1.93$. This threshold is then fixed for all downstream datasets and attacks. 

\subsection{Ablation Study on Target Prompt $p$}
\label{app:target}

\begin{wraptable}{r}{0.45\linewidth}
\vspace{-0.4cm}
\centering
\caption{Target-prompt sensitivity.}
\label{tab:target_prompt}
\setlength{\tabcolsep}{1.7pt}
\renewcommand{\arraystretch}{0.9}
\setlength{\belowcaptionskip}{5pt}
\begin{tabular}{lcc}
\toprule
\textbf{Target prompt} & \textbf{Clean} & \textbf{Robustness} \\
\midrule
a white teapot  & 61.1 & 52.3 \\
a red sports car & 61.1 & 52.3 \\
a wooden chair  & 61.0 & 52.3 \\
a cute cat       & 61.1 & 52.1 \\
\bottomrule
\end{tabular}
\vspace{-0.2cm}
\end{wraptable}
Recall that the target prompt specifies the observable output direction associated with the implanted probe. We therefore investigate whether BaP is sensitive to its semantic content.  Table~\ref{tab:target_prompt} examines the sensitivity of BaP to four semantically distinct target prompts. The results remain highly consistent in terms of clean accuracy and robustness. This stability indicates that the performance of BaP does not rely on the specific semantic content of a backdoor target.

\section{Detailed Prompt for GPTScore}
\label{app:prompt}

\begin{figure}[h]
\centering
\includegraphics[width=\linewidth]{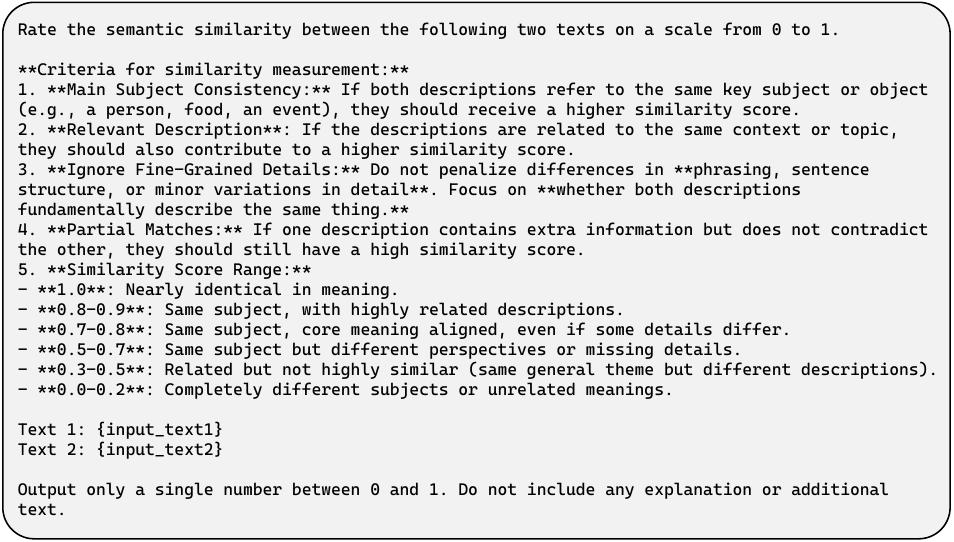}
\caption{The system prompt for computing GPTScore.}
\label{fig:prompt}
\end{figure}

Followed by~\citep{li2025a}, we compute the captioning performance via GPTScore. The detailed prompt for GPTScore is provided in Fig.~\ref{fig:prompt}.

\section{Detailed Results on CLIP-B/32 and CLIP-L/14}
\label{app:transfer_results}

Tables~\ref{tab:clip_b32_comparison} and
\ref{tab:clip_l14_comparison} report the per-dataset results across
16 datasets for CLIP ViT-B/32 and ViT-L/14, respectively. These results
extend the category-level comparison in Table~\ref{tab:backbone_transfer}
and further demonstrate the effectiveness of BaP across different CLIP backbones.

\section{Detection ROC of BaP}
\label{app:auc}

\begin{figure}[t]
\centering
\includegraphics[width=\linewidth]{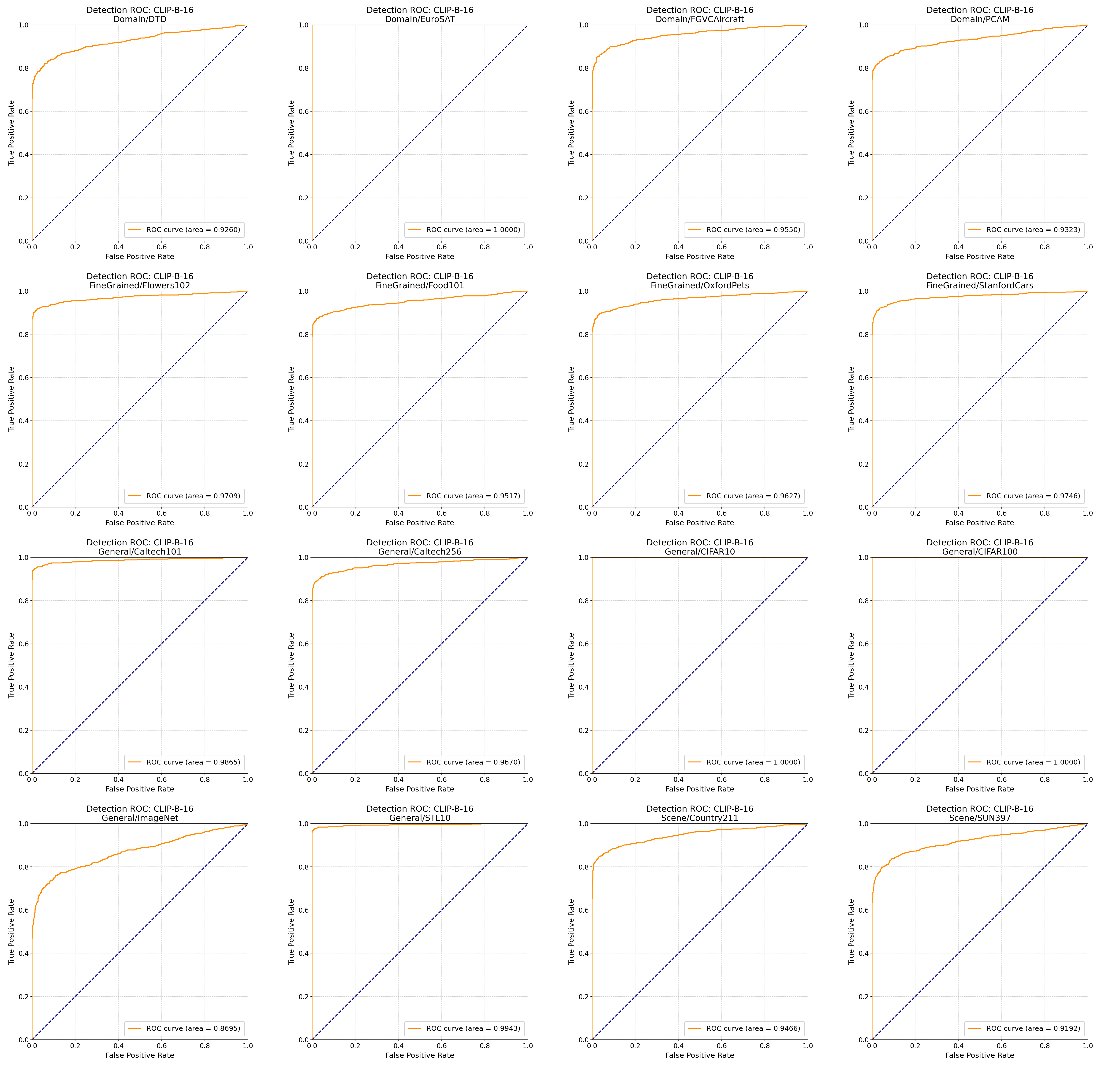}
\caption{ROC curves of CLIP-B/16 for adversarial sample detection on 16 datasets.}
\label{fig:auc_curves_b16}
\end{figure}

\begin{figure}[t]
\centering
\includegraphics[width=\linewidth]{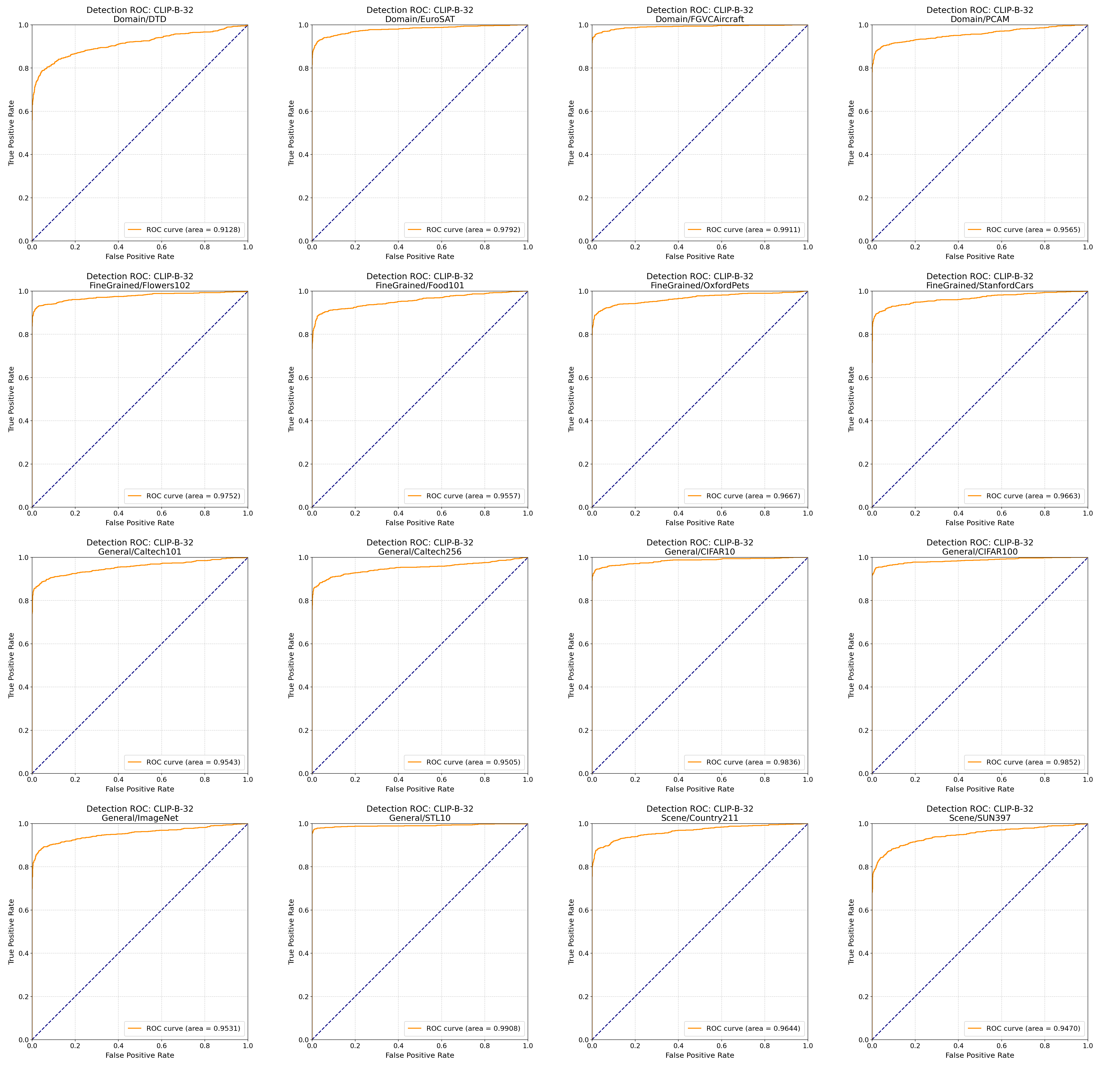}
\caption{ROC curves of CLIP-B/32 for adversarial sample detection on 16 datasets.}
\label{fig:auc_curves_b32}
\end{figure}

\begin{figure}[t]
\centering
\includegraphics[width=\linewidth]{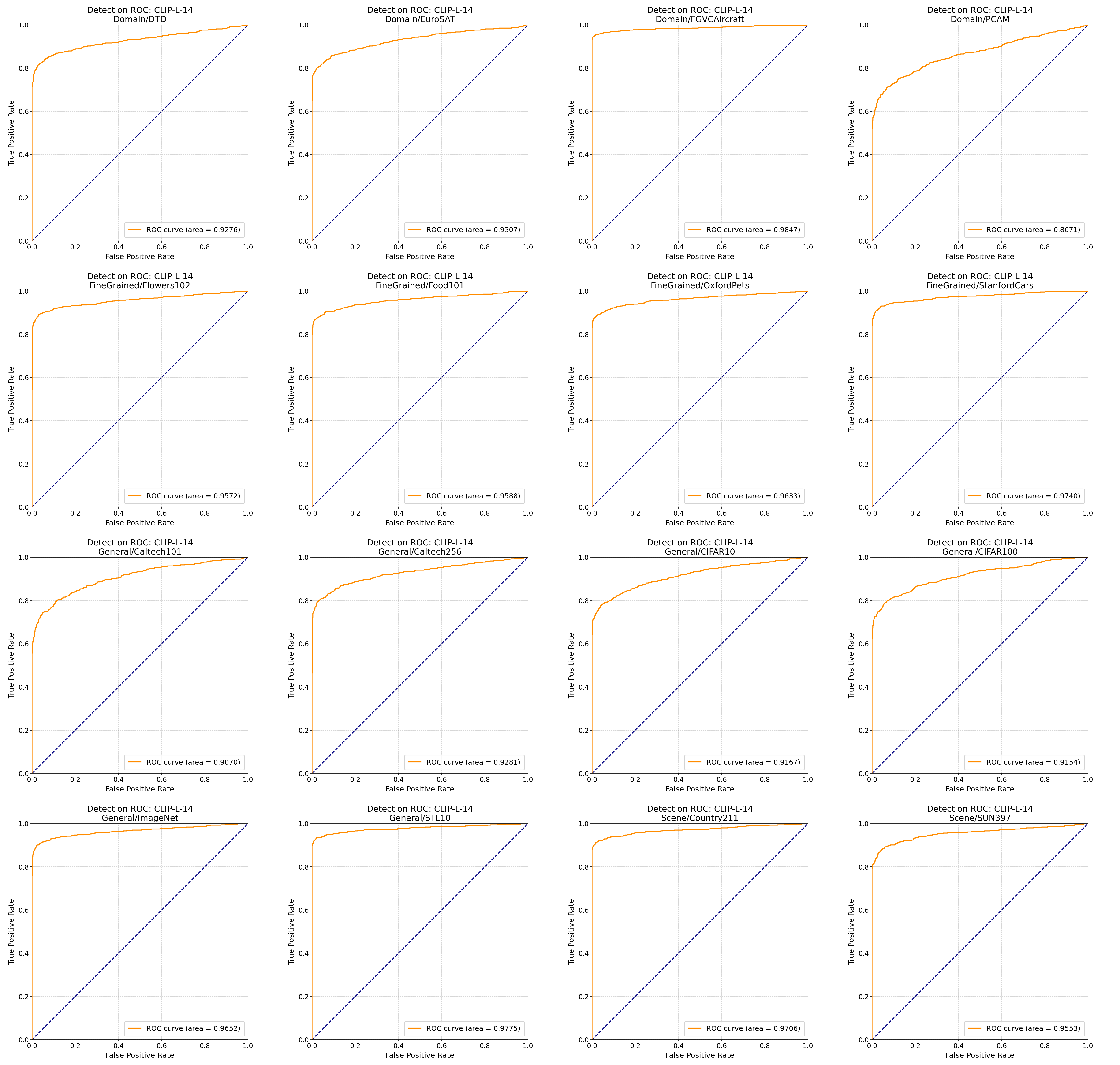}
\caption{ROC curves of CLIP-L/14 for adversarial sample detection on 16 datasets.}
\label{fig:auc_curves_l14}
\end{figure}

\begin{figure}[t]
\centering

\begin{subfigure}{0.38\linewidth}
    \centering
    \includegraphics[width=\linewidth]{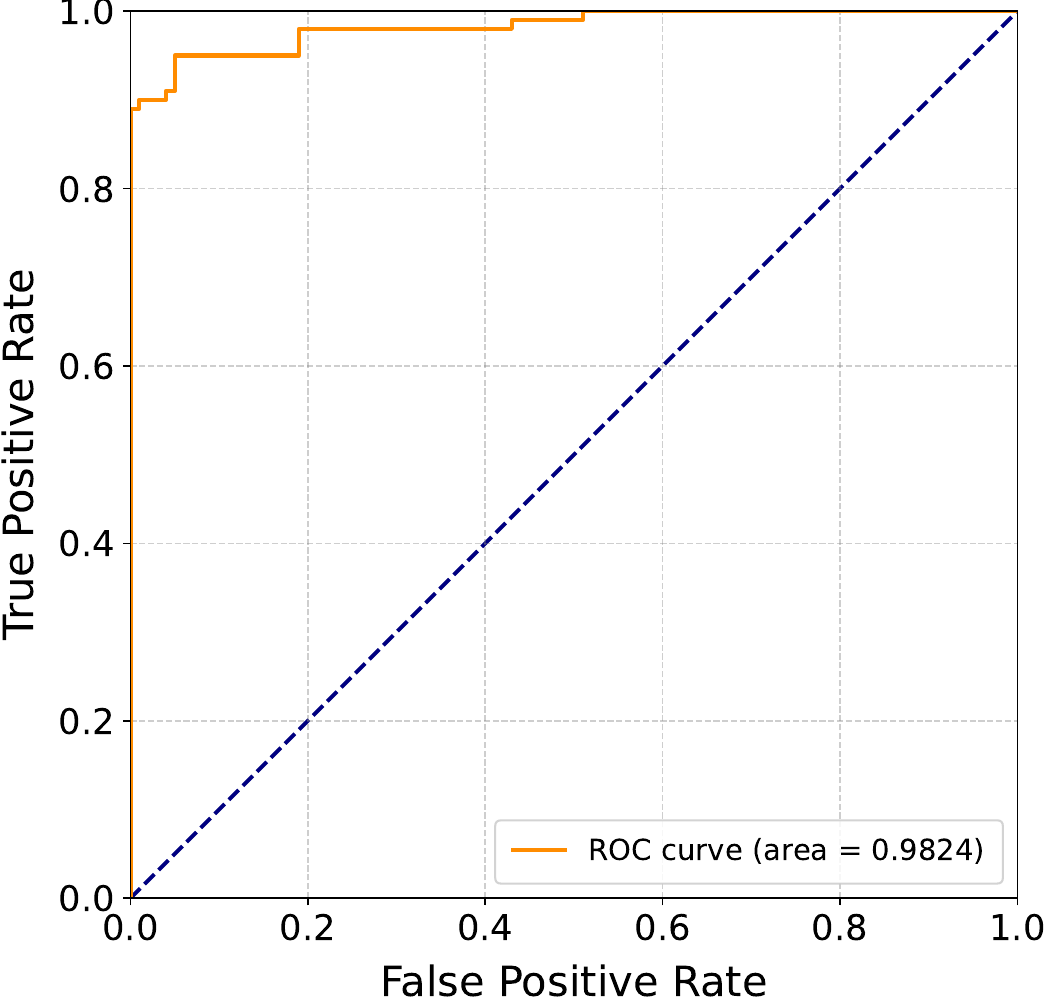}
    \caption{FOA-Attack.}
    \label{fig:foa_attack_roc_auc}
\end{subfigure}
\begin{subfigure}{0.38\linewidth}
    \centering
    \includegraphics[width=\linewidth]{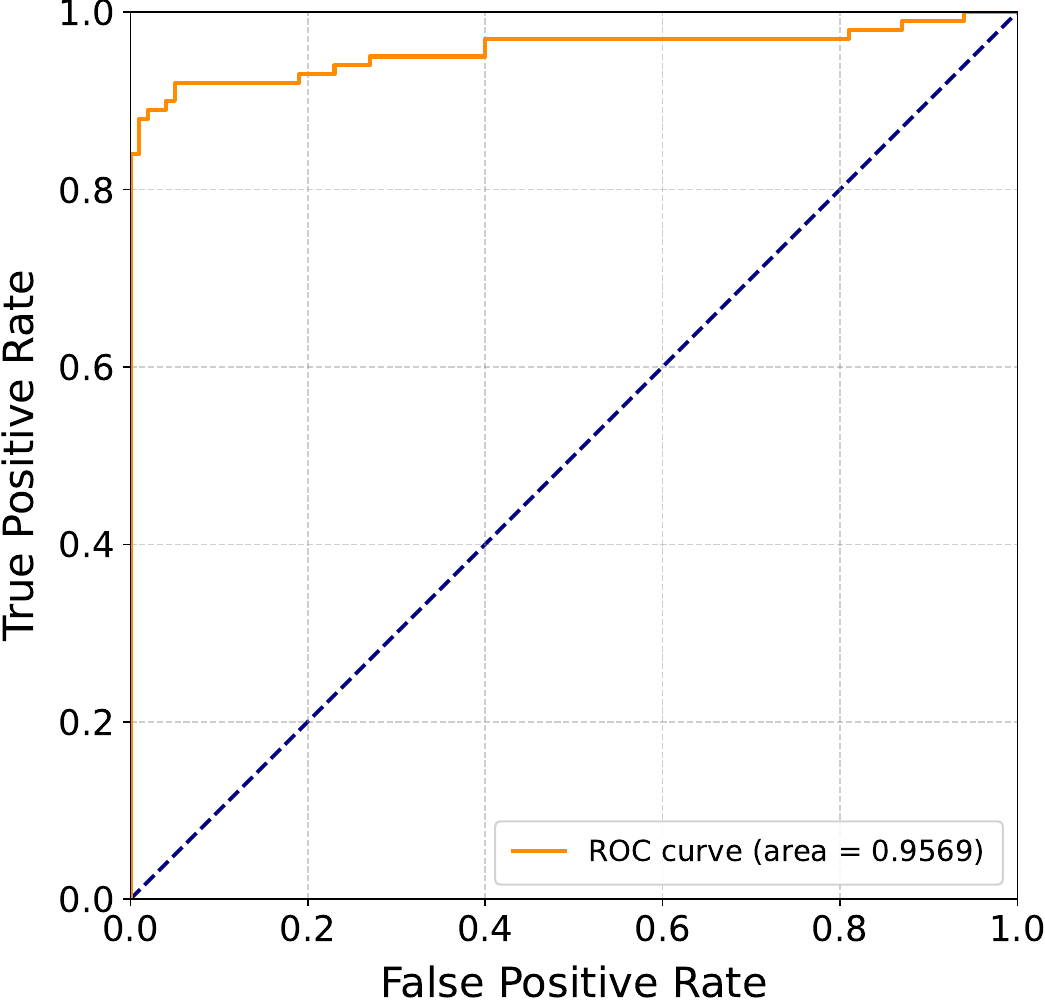}
    \caption{M-Attack}
    \label{fig:m_attack_roc_auc}
\end{subfigure}

\caption{ROC curves of CLIP-L/14@336 for LVLM attacks.}
\label{fig:LVLMs_attack}
\end{figure}

The ROC curves across different backbones are provided in Fig.~\ref{fig:auc_curves_b16}, Fig.~\ref{fig:auc_curves_b32}, Fig.~\ref{fig:auc_curves_l14} and Fig.~\ref{fig:LVLMs_attack}. BaP achieves consistently high AUC values across backbones and attacks, indicating that its internal probe effectively distinguishes adversarial inputs from clean ones.

\begin{table*}[t]
\centering
\caption{Performance comparison across different dataset types and method categories on \textbf{CLIP-B/32}. $\Delta$ reports the difference between BaP and the original CLIP.}
\label{tab:clip_b32_comparison}
\setlength{\tabcolsep}{2.0pt}
\renewcommand{\arraystretch}{1.08}
\resizebox{\textwidth}{!}{%
\begin{tabular}{ll||cc|cc|cc|cc|cc|cc|cc|cc|cc||cc}
\toprule
\multicolumn{2}{c}{\textbf{Dataset}}
& \multicolumn{2}{c}{\textbf{Original}}
& \multicolumn{16}{c}{\textbf{Test-Time Defense}}
& \multicolumn{2}{c}{\textbf{$\Delta$}} \\
\cmidrule(lr){3-4}
\cmidrule(lr){5-20}
\cmidrule(lr){21-22}
\multicolumn{2}{c}{}
& \multicolumn{2}{c}{\textbf{CLIP}}
& \multicolumn{2}{c}{\textbf{R-TPT}}
& \multicolumn{2}{c}{\textbf{LPF}}
& \multicolumn{2}{c}{\textbf{HD}}
& \multicolumn{2}{c}{\textbf{Anti-Adv}}
& \multicolumn{2}{c}{\textbf{TTE}}
& \multicolumn{2}{c}{\textbf{TTC}}
& \multicolumn{2}{c}{\textbf{ET3}}
& \multicolumn{2}{c}{\textbf{BaP (Ours)}}
& \multicolumn{2}{c}{} \\
\textbf{Type} & \textbf{Name}
& Clean & Rob. & Clean & Rob. & Clean & Rob. & Clean & Rob.
& Clean & Rob. & Clean & Rob. & Clean & Rob. & Clean & Rob.
& Clean & Rob. & Clean & Rob. \\
\midrule

& ImageNet
& \textcolor{clipgray}{57.8} & \textcolor{clipgray}{0.2}
& 57.4 & 30.2 & 52.0 & 17.4 & 55.2 & 3.5 & 55.7 & 10.8
& \textbf{61.3} & 24.1 & 44.0 & 24.6 & 51.4 & 12.4 & 53.4 & \textbf{43.6}
& \textcolor{blue!80}{-4.4} & \textcolor{red!80}{+43.4} \\
\rowcolor{gray!10}
& CIFAR10
& \textcolor{clipgray}{86.1} & \textcolor{clipgray}{0.6}
& 76.7 & 35.1 & 84.9 & 27.2 & 86.5 & 4.2 & 84.4 & 41.8
& 86.0 & 33.7 & \textbf{87.6} & 43.3 & 74.1 & 30.4 & 87.1 & \textbf{63.1}
& \textcolor{red!80}{+1.0} & \textcolor{red!80}{+62.5} \\
& CIFAR100
& \textcolor{clipgray}{57.2} & \textcolor{clipgray}{0.3}
& 41.0 & 14.7 & 56.0 & 10.3 & \textbf{61.6} & 4.2 & 53.5 & 20.9
& 58.8 & 15.5 & 58.8 & 19.1 & 43.5 & 14.8 & 55.5 & \textbf{35.5}
& \textcolor{blue!80}{-1.7} & \textcolor{red!80}{+35.2} \\
\rowcolor{gray!10}
& STL10
& \textcolor{clipgray}{96.2} & \textcolor{clipgray}{12.3}
& 96.5 & 78.5 & 96.0 & 67.2 & 95.2 & 32.6 & 95.3 & 67.8
& \textbf{97.4} & 84.8 & 96.5 & 72.5 & 90.7 & 61.1 & 95.6 & \textbf{92.5}
& \textcolor{blue!80}{-0.6} & \textcolor{red!80}{+80.2} \\
& Caltech101
& \textcolor{clipgray}{82.3} & \textcolor{clipgray}{6.3}
& \textbf{84.9} & 29.4 & 81.0 & 56.7 & 81.9 & 29.1 & 83.5 & 50.1
& 83.2 & 65.7 & 82.5 & 52.1 & 79.6 & 47.4 & 79.9 & \textbf{71.8}
& \textcolor{blue!80}{-2.4} & \textcolor{red!80}{+65.5} \\
\rowcolor{gray!10}
\multirow{-6}{*}{\rotatebox[origin=c]{90}{General}} & Caltech256
& \textcolor{clipgray}{80.3} & \textcolor{clipgray}{4.1}
& \textbf{80.8} & 63.5 & 78.6 & 49.3 & 78.8 & 21.7 & 79.2 & 43.1
& 78.5 & 60.0 & 77.3 & 49.2 & 77.2 & 40.3 & 77.0 & \textbf{67.9}
& \textcolor{blue!80}{-3.3} & \textcolor{red!80}{+63.8} \\
\midrule

& OxfordPets
& \textcolor{clipgray}{84.2} & \textcolor{clipgray}{0.2}
& \textbf{84.2} & 60.2 & 74.6 & 25.2 & \textbf{84.2} & 6.4 & 83.6 & 19.9
& 81.7 & 24.8 & 82.2 & 30.7 & 75.2 & 24.3 & 79.1 & \textbf{66.4}
& \textcolor{blue!80}{-5.1} & \textcolor{red!80}{+66.2} \\
\rowcolor{gray!10}
& Flowers102
& \textcolor{clipgray}{62.7} & \textcolor{clipgray}{0.8}
& 59.2 & 35.2 & 57.8 & 25.0 & 59.8 & 6.2 & 60.9 & 14.2
& 61.5 & 14.2 & \textbf{61.9} & 26.4 & 56.5 & 18.9 & 60.3 & \textbf{50.0}
& \textcolor{blue!80}{-2.4} & \textcolor{red!80}{+49.2} \\
& Food101
& \textcolor{clipgray}{80.4} & \textcolor{clipgray}{0.1}
& \textbf{82.1} & 57.1 & 71.0 & 20.3 & 79.7 & 3.6 & 78.5 & 11.4
& 79.9 & 46.0 & 72.7 & 36.5 & 73.7 & 13.1 & 77.1 & \textbf{60.4}
& \textcolor{blue!80}{-3.3} & \textcolor{red!80}{+60.3} \\
\rowcolor{gray!10}
\multirow{-4}{*}{\rotatebox[origin=c]{90}{Fine-G}} & StanfordCars
& \textcolor{clipgray}{59.9} & \textcolor{clipgray}{0.0}
& \textbf{60.2} & 28.1 & 44.4 & 5.1 & 50.5 & 3.2 & 58.1 & 4.9
& 51.2 & 26.6 & 49.0 & 16.3 & 42.7 & 6.1 & 47.3 & \textbf{42.7}
& \textcolor{blue!80}{-12.6} & \textcolor{red!80}{+42.7} \\
\midrule

& SUN397
& \textcolor{clipgray}{61.9} & \textcolor{clipgray}{0.8}
& \textbf{63.1} & \textbf{54.2} & 59.3 & 21.3 & 58.1 & 6.1 & 61.7 & 13.2
& 62.2 & 18.0 & 53.1 & 31.2 & 58.8 & 15.7 & 61.0 & 52.7
& \textcolor{blue!80}{-0.9} & \textcolor{red!80}{+51.9} \\
\rowcolor{gray!10}
\multirow{-2}{*}{\rotatebox[origin=c]{90}{Scene}} & Country211
& \textcolor{clipgray}{15.7} & \textcolor{clipgray}{0.0}
& 14.3 & 5.0 & 13.1 & 1.2 & 12.1 & 0.0 & 12.9 & 0.8
& \textbf{16.3} & 1.2 & 12.6 & 3.7 & 11.2 & 1.8 & 14.0 & \textbf{9.3}
& \textcolor{blue!80}{-1.7} & \textcolor{red!80}{+9.3} \\
\midrule

& FGVCAircraft
& \textcolor{clipgray}{17.3} & \textcolor{clipgray}{0.0}
& 19.7 & 13.6 & 16.7 & 2.0 & 15.7 & 1.1 & 14.3 & 2.0
& \textbf{19.8} & 5.1 & 11.4 & 9.2 & 14.1 & 1.8 & 17.1 & \textbf{19.8}
& \textcolor{blue!80}{-0.2} & \textcolor{red!80}{+19.8} \\
\rowcolor{gray!10}
& EuroSAT
& \textcolor{clipgray}{34.2} & \textcolor{clipgray}{0.0}
& 27.7 & 21.4 & 29.5 & 1.5 & \textbf{35.3} & 2.1 & 28.7 & 13.7
& 28.8 & 10.2 & 33.6 & 13.9 & 26.9 & 12.9 & 34.2 & \textbf{32.1}
& \textcolor{red!80}{+0.0} & \textcolor{red!80}{+32.1} \\
& DTD
& \textcolor{clipgray}{43.3} & \textcolor{clipgray}{1.7}
& 39.6 & \textbf{32.6} & 39.9 & 18.0 & 40.3 & 11.4 & 40.9 & 15.6
& \textbf{44.2} & 23.2 & 42.9 & 22.1 & 35.6 & 18.7 & 38.8 & 28.9
& \textcolor{blue!80}{-4.5} & \textcolor{red!80}{+27.2} \\
\rowcolor{gray!10}
\multirow{-4}{*}{\rotatebox[origin=c]{90}{Domain}} & PCAM
& \textcolor{clipgray}{48.9} & \textcolor{clipgray}{24.8}
& \textbf{55.4} & 41.2 & 48.7 & 48.2 & 49.0 & 39.8 & 48.8 & 48.3
& 54.5 & 38.7 & 48.8 & 44.2 & 48.6 & 48.4 & 49.1 & \textbf{49.9}
& \textcolor{red!80}{+0.2} & \textcolor{red!80}{+25.1} \\
\midrule

\rowcolor{lightblue}
All & Avg.
& \textcolor{clipgray}{60.5} & \textcolor{clipgray}{3.3}
& 57.9 & 37.6 & 56.5 & 24.7 & \textbf{59.0} & 11.0 & 58.8 & 23.7
& 57.2 & 30.7 & 57.2 & 30.9 & 53.7 & 23.0 & 57.9 & \textbf{49.2}
& \textcolor{blue!80}{-2.6} & \textcolor{red!80}{+45.9} \\
\bottomrule
\end{tabular}}
\end{table*}

\begin{table*}[t]
\renewcommand{\arraystretch}{1.2}
\centering
\setlength{\tabcolsep}{2.5pt}
\caption{Performance comparison across different dataset types and method categories on \textbf{CLIP-L/14}. $\Delta$ reports the difference between BaP and the original CLIP.}
\label{tab:clip_l14_comparison}
\resizebox{\textwidth}{!}{%
\begin{tabular}{ll||cc|cc|cc|cc|cc|cc|cc|cc|cc||cc}
\toprule
\multicolumn{2}{c}{\textbf{Dataset}}
& \multicolumn{2}{c}{\textbf{Original}}
& \multicolumn{16}{c}{\textbf{Test-Time Defense}}
& \multicolumn{2}{c}{\textbf{$\Delta$}} \\
\cmidrule(lr){3-4}
\cmidrule(lr){5-20}
\cmidrule(lr){21-22}
\multicolumn{2}{c}{}
& \multicolumn{2}{c}{\textbf{CLIP}}
& \multicolumn{2}{c}{\textbf{R-TPT}}
& \multicolumn{2}{c}{\textbf{LPF}}
& \multicolumn{2}{c}{\textbf{HD}}
& \multicolumn{2}{c}{\textbf{Anti-Adv}}
& \multicolumn{2}{c}{\textbf{TTE}}
& \multicolumn{2}{c}{\textbf{TTC}}
& \multicolumn{2}{c}{\textbf{ET3}}
& \multicolumn{2}{c}{\textbf{BaP (Ours)}}
& \multicolumn{2}{c}{} \\
\textbf{Type} & \textbf{Name}
& Clean & Rob. & Clean & Rob. & Clean & Rob. & Clean & Rob.
& Clean & Rob. & Clean & Rob. & Clean & Rob. & Clean & Rob.
& Clean & Rob. & Clean & Rob. \\
\midrule

& ImageNet
& \textcolor{clipgray}{68.1} & \textcolor{clipgray}{0.6}
& 71.2 & 60.0 & 64.2 & 43.2 & 66.4 & 10.1 & 68.0 & 37.1
& \textbf{71.5} & 32.4 & 54.7 & 35.8 & 66.3 & 16.1 & 65.9 & \textbf{60.7}
& \textcolor{blue!80}{-2.2} & \textcolor{red!80}{+60.1} \\
\rowcolor{gray!10}
& CIFAR10
& \textcolor{clipgray}{93.4} & \textcolor{clipgray}{1.1}
& 90.2 & \textbf{82.1} & \textbf{94.3} & 71.0 & 92.1 & 40.5 & 89.3 & 78.5
& 92.3 & 47.0 & 94.1 & 18.4 & 86.7 & 45.9 & 93.2 & 53.7
& \textcolor{blue!80}{-0.2} & \textcolor{red!80}{+52.6} \\
& CIFAR100
& \textcolor{clipgray}{65.0} & \textcolor{clipgray}{0.1}
& 65.7 & \textbf{58.3} & 72.8 & 40.3 & 67.9 & 26.7 & 64.7 & 51.8
& \textbf{72.9} & 37.9 & 71.0 & 7.0 & 62.7 & 27.7 & 62.3 & 28.6
& \textcolor{blue!80}{-2.7} & \textcolor{red!80}{+28.5} \\
\rowcolor{gray!10}
& STL10
& \textcolor{clipgray}{99.4} & \textcolor{clipgray}{12.5}
& 98.8 & \textbf{93.4} & 99.2 & 91.7 & 98.6 & 67.8 & 99.0 & 93.3
& 98.6 & 88.9 & \textbf{99.4} & 53.2 & 98.2 & 72.4 & 99.1 & 92.9
& \textcolor{blue!80}{-0.3} & \textcolor{red!80}{+80.4} \\
& Caltech101
& \textcolor{clipgray}{85.3} & \textcolor{clipgray}{5.0}
& 89.0 & \textbf{81.6} & 86.3 & 77.5 & 84.7 & 45.2 & 86.0 & 72.6
& \textbf{91.2} & 73.5 & 81.6 & 41.6 & 84.5 & 50.5 & 84.2 & 65.7
& \textcolor{blue!80}{-1.1} & \textcolor{red!80}{+60.7} \\
\rowcolor{gray!10}
\multirow{-6}{*}{\rotatebox[origin=c]{90}{General}} & Caltech256
& \textcolor{clipgray}{88.4} & \textcolor{clipgray}{4.6}
& \textbf{90.5} & \textbf{85.4} & 87.4 & 77.9 & 86.2 & 39.5 & 88.0 & 74.3
& 89.9 & 71.3 & 82.0 & 44.0 & 88.0 & 47.2 & 87.6 & 72.5
& \textcolor{blue!80}{-0.8} & \textcolor{red!80}{+67.9} \\
\midrule

& OxfordPets
& \textcolor{clipgray}{89.9} & \textcolor{clipgray}{0.2}
& \textbf{94.2} & \textbf{81.4} & 88.3 & 61.7 & 90.5 & 13.9 & 91.7 & 61.1
& 90.1 & 36.0 & 84.8 & 46.8 & 87.6 & 24.4 & 88.7 & 77.6
& \textcolor{blue!80}{-1.2} & \textcolor{red!80}{+77.4} \\
\rowcolor{gray!10}
& Flowers102
& \textcolor{clipgray}{72.4} & \textcolor{clipgray}{0.7}
& 71.5 & 60.8 & 72.5 & 50.1 & 70.9 & 10.3 & \textbf{73.5} & 46.3
& 73.0 & 34.9 & 66.0 & 33.6 & 70.4 & 16.4 & 71.2 & \textbf{64.6}
& \textcolor{blue!80}{-1.2} & \textcolor{red!80}{+63.9} \\
& Food101
& \textcolor{clipgray}{90.0} & \textcolor{clipgray}{0.1}
& \textbf{90.6} & \textbf{79.3} & 86.8 & 61.2 & 88.7 & 7.0 & 87.7 & 53.1
& 89.8 & 36.8 & 68.3 & 47.9 & 86.5 & 16.3 & 87.3 & 75.0
& \textcolor{blue!80}{-2.7} & \textcolor{red!80}{+74.9} \\
\rowcolor{gray!10}
\multirow{-4}{*}{\rotatebox[origin=c]{90}{Fine-G}} & StanfordCars
& \textcolor{clipgray}{73.0} & \textcolor{clipgray}{0.2}
& \textbf{77.5} & 59.8 & 67.6 & 33.9 & 68.8 & 5.8 & 76.8 & 31.6
& 73.4 & 21.4 & 61.9 & 30.3 & 67.2 & 8.4 & 71.8 & \textbf{63.3}
& \textcolor{blue!80}{-1.2} & \textcolor{red!80}{+63.1} \\
\midrule

& SUN397
& \textcolor{clipgray}{68.2} & \textcolor{clipgray}{0.5}
& \textbf{69.9} & 61.5 & 65.6 & 45.3 & 64.9 & 11.5 & 68.5 & 38.0
& 69.4 & 22.1 & 56.4 & 33.0 & 66.4 & 15.7 & 65.5 & \textbf{61.9}
& \textcolor{blue!80}{-2.7} & \textcolor{red!80}{+61.4} \\
\rowcolor{gray!10}
\multirow{-2}{*}{\rotatebox[origin=c]{90}{Scene}} & Country211
& \textcolor{clipgray}{23.8} & \textcolor{clipgray}{0.0}
& 24.2 & 13.4 & 22.8 & 7.1 & 21.1 & 0.8 & 21.9 & 6.0
& \textbf{26.2} & 2.3 & 13.8 & 6.3 & 20.3 & 1.9 & 23.3 & \textbf{19.4}
& \textcolor{blue!80}{-0.5} & \textcolor{red!80}{+19.4} \\
\midrule

& FGVCAircraft
& \textcolor{clipgray}{28.1} & \textcolor{clipgray}{0.0}
& \textbf{33.2} & 20.7 & 26.1 & 11.1 & 26.5 & 1.4 & 26.5 & 10.7
& 30.2 & 5.3 & 23.4 & 12.1 & 27.3 & 1.5 & 26.5 & \textbf{29.2}
& \textcolor{blue!80}{-1.6} & \textcolor{red!80}{+29.2} \\
\rowcolor{gray!10}
& EuroSAT
& \textcolor{clipgray}{54.8} & \textcolor{clipgray}{0.1}
& 38.8 & \textbf{33.6} & \textbf{53.2} & 19.0 & 47.5 & 11.5 & 46.9 & 30.4
& 41.1 & 12.0 & 52.1 & 7.7 & 45.3 & 16.4 & 52.3 & 24.8
& \textcolor{blue!80}{-2.5} & \textcolor{red!80}{+24.7} \\
& DTD
& \textcolor{clipgray}{53.0} & \textcolor{clipgray}{0.7}
& \textbf{53.8} & \textbf{44.4} & 48.9 & 35.2 & 52.3 & 15.1 & 50.3 & 33.5
& 52.5 & 26.6 & 42.4 & 25.6 & 50.2 & 23.1 & 51.2 & 43.0
& \textcolor{blue!80}{-1.8} & \textcolor{red!80}{+42.3} \\
\rowcolor{gray!10}
\multirow{-4}{*}{\rotatebox[origin=c]{90}{Domain}} & PCAM
& \textcolor{clipgray}{49.6} & \textcolor{clipgray}{0.2}
& 43.6 & 50.4 & 49.9 & 48.8 & 50.4 & 35.4 & 50.2 & 49.1
& \textbf{50.8} & \textbf{50.6} & 50.1 & 16.6 & 50.1 & 47.0 & 49.5 & 44.9
& \textcolor{blue!80}{-0.1} & \textcolor{red!80}{+44.7} \\
\midrule

\rowcolor{lightblue}
All & Avg.
& \textcolor{clipgray}{68.9} & \textcolor{clipgray}{1.7}
& 68.9 & \textbf{60.4} & 67.9 & 48.0 & 67.3 & 21.4 & 68.1 & 54.2
& \textbf{69.6} & 37.4 & 62.6 & 28.7 & 66.1 & 26.9 & 67.5 & 54.9
& \textcolor{blue!80}{-1.4} & \textcolor{red!80}{+53.2} \\
\bottomrule
\end{tabular}}
\end{table*}

\end{document}